\documentclass[letterpaper, 10 pt, conference]{Format/ieeeconf}

\usepackage[only-used=true, single=false]{acro}
\usepackage[ruled,vlined,linesnumbered]{algorithm2e}
\usepackage{amsfonts}
\usepackage{amsmath, amssymb}
\usepackage{array}
\usepackage{bm}
\usepackage{cite}
\usepackage{color}
\usepackage{float}
\usepackage[utf8]{inputenc} 
\usepackage[T1]{fontenc}    
\usepackage{graphicx}
\usepackage{subfig}
\usepackage{url}
\usepackage{mathtools}

\usepackage{multirow}
\usepackage{booktabs}

\usepackage{soul,xcolor}

\IEEEoverridecommandlockouts
\title{\LARGE \bf Learning Generalisable Coupling Terms for Obstacle Avoidance \\ via Low-dimensional Geometric Descriptors}
\author{{\`E}ric Pairet$^{*}$, Paola Ard{\'o}n, Michael Mistry, and Yvan Petillot
\thanks{Authors are with the Edinburgh Centre for Robotics at University of Edinburgh and Heriot-Watt University (UK). ${}^{*}${\tt\small eric.pairet@ed.ac.uk}}%
}

\newcommand*{\sref}[1]{Section~\ref{#1}}

\newcommand*{\eref}[1]{(\ref{#1})}

\newcommand*{\fref}[1]{Figure~\ref{#1}}
\newcommand*{\tref}[1]{Table~\ref{#1}}

\DeclareMathOperator{\sign}{sign}
\newcommand{\minus}{\text{-}}

\newcommand{\real}{\mathbb{R}}

\newcommand{\rotmat}{\mathrm{SO}}

\newcommand{\vk}{\mathbf{k}}

\newcommand{\pos}{x}
\newcommand{\vel}{\dot{x}}

\newcommand{\vpos}{\mathbf{x}}
\newcommand{\vvel}{\mathbf{\dot{x}}}

\newcommand{\forcingterm}{f(\boldsymbol{\cdot})}
\newcommand{\couplingterm}{C(\boldsymbol{\cdot})}
\newcommand{\couplingtermOA}{C_{\mbox{\tiny OA}}(\boldsymbol{\cdot})}
\newcommand{\couplingtermHG}{C_{\mbox{\tiny HG}}(\boldsymbol{\cdot})}

\newcommand{\couplingtermOAi}{C_{\mbox{\tiny OA}}^{\,i}(\boldsymbol{\cdot})}
\newcommand{\couplingtermHGi}{C_{\mbox{\tiny HG}}^{\,i}(\boldsymbol{\cdot})}

\newcommand{\newgamma}{\alpha}
\newcommand{\newbeta}{\psi}
\newcommand{\newkappa}{\kappa}

\newcommand{\projectedwidth}{\lambda_{1}^{\prime}}
\newcommand{\projectedheight}{\lambda_{2}^{\prime}}

\newcommand{\vprojectedgeometry}{\boldsymbol{\lambda}^{\prime}}
\newcommand{\clearance}{\Delta}

\newcommand{\inputvector}{\mathbf{h}}

\newcommand{\pplane}{\textsc{P}}
\newcommand{\yzplane}{\textsc{YZ}}

\newcommand{\cost}{\eta}

\usepackage[normalem]{ulem}
\definecolor{orange}{RGB}{255,128,0}
\definecolor{blue}{RGB}{0,128,255}

\DeclareAcronym{1D}{
  short = 1D,
  long  = one-dimensional
}
\DeclareAcronym{2D}{
  short = 2D,
  long  = two-dimensional
}
\DeclareAcronym{3D}{
  short = 3D,
  long  = three-dimensional
}
\DeclareAcronym{AI}{
  short = AI,
  long  = artificial intelligence
}
\DeclareAcronym{CAN}{
  short = CAN,
  long  = controller area network
}
\DeclareAcronym{CF}{
  short = CF,
  long  = coupling force
}
\DeclareAcronym{CMP}{
  short = CMP,
  long  = compliant movement primitive
}
\DeclareAcronym{DMP}{
  short = DMP,
  long  = dynamic movement primitive
}
\DeclareAcronym{DoF}{
  short = DoF,
  long  = degree of freedom,
  long-plural-form = degrees of freedom
}
\DeclareAcronym{DS}{
  short = DS,
  long  = dynamical system
}
\DeclareAcronym{ECR}{
  short = ECR,
  long  = Edinburgh Centre for Robotics
}
\DeclareAcronym{EM}{
  short = EM,
  long  = expectation-maximisation
}
\DeclareAcronym{FK}{
  short = FK,
  long  = forwad kinematics
}
\DeclareAcronym{GMM}{
  short = GMM,
  long  = Gaussian mixture model
}
\DeclareAcronym{GMR}{
  short = GMR,
  long  = Gaussian mixture regression
}
\DeclareAcronym{GPR}{
  short = GPR,
  long  = Gaussian process regression
}
\DeclareAcronym{HMM}{
  short = HMM,
  long  = hidden Markov model
}
\DeclareAcronym{HRI}{
  short = HRI,
  long  = human-robot interaction
}
\DeclareAcronym{HSMM}{
  short = HSMM,
  long  = hidden semi-Markov model
}
\DeclareAcronym{KL}{
  short = KL,
  long  = Kullback-Leibler
}
\DeclareAcronym{IGMM}{
  short = IGMM,
  long  = infinite Gaussian mixture model
}
\DeclareAcronym{IIT}{
  short = IIT,
  long  = Italian Institute of Technology
}
\DeclareAcronym{IK}{
  short = IK,
  long  = inverse kinematics
}
\DeclareAcronym{ILC}{
  short = ILC,
  long  = iterative learning control
}
\DeclareAcronym{IR}{
  short = IR,
  long  = infra-red
}
\DeclareAcronym{LbD}{
  short = LbD,
  long  = learning by demonstration
}
\DeclareAcronym{LED}{
  short = LED,
  long  = light-emitting diode
}
\DeclareAcronym{LMS}{
  short = LMS,
  long  = least mean squares
}
\DeclareAcronym{LS}{
  short = LS,
  long  = linear square
}
\DeclareAcronym{LWPR}{
  short = LWPR,
  long  = locally weighted projection regression
}
\DeclareAcronym{LWR}{
  short = LWR,
  long  = locally weighted regression
}
\DeclareAcronym{MTR}{
  short = MTR,
  long  = multiple target regression
}
\DeclareAcronym{NMSE}{
  short = NMSE,
  long  = normalised mean squared error
}
\DeclareAcronym{NN}{
  short = NN,
  long  = neural network
}
\DeclareAcronym{OROCOS}{
  short = OROCOS,
  long  = open robot control software
}
\DeclareAcronym{PD}{
  short = PD,
  long  = proportional-derivative
}
\DeclareAcronym{RBF}{
  short = RBF,
  long  = radial basis function
}
\DeclareAcronym{RC}{
  short = RC,
  long  = regressor chain
}
\DeclareAcronym{ReLu}{
  short = ReLu,
  long  = rectified linear unit
}
\DeclareAcronym{RFWR}{
  short = RFWR,
  long  = receptive field weighted regression
}
\DeclareAcronym{RL}{
  short = RL,
  long  = reinforcement learning
}
\DeclareAcronym{ROS}{
  short = ROS,
  long  = robot operating system
}
\DeclareAcronym{STR}{
  short = STR,
  long  = single target regression
}
\DeclareAcronym{WP}{
  short = WP,
  long  = work package
}
\DeclareAcronym{YARP}{
  short = YARP,
  long  = yet another robotic platform
}

\begin{document}
    \maketitle
    \thispagestyle{empty}
    \pagestyle{empty}

    \begin{abstract}
    Unforeseen events are frequent in the real-world environments where robots are expected to assist, raising the need for fast replanning of the policy in execution to guarantee the system and environment safety. Inspired by human behavioural studies of obstacle avoidance and route selection, this paper presents a hierarchical framework which generates reactive yet bounded obstacle avoidance behaviours through a multi-layered analysis. The framework leverages the strengths of learning techniques and the versatility of \aclp{DMP} to efficiently unify perception, decision, and action levels via low-dimensional geometric descriptors of the environment. 
    Experimental evaluation on synthetic environments and a real anthropomorphic manipulator proves that the robustness and generalisation capabilities of the proposed approach regardless of the obstacle avoidance scenario makes it suitable for robotic systems in real-world environments.
\end{abstract}
    \section{INTRODUCTION} \label{sec:introduction}
    Robust reactive behaviours are essential to ensure the safety of robots operating in unstructured environments. For instance, the on-going pick-and-place policy of a robotic system sorting and storing items in a home environment might be interrupted by the sudden appearance of an obstacle in the middle of a pre-planned trajectory. In this scenario, the robot must be able to modulate its behaviour online to succeed in its task while providing some safety guarantees. Given the expertise of humans in dealing with these conditions, it is natural to adopt human behaviour for robotic control.

    Human behavioural studies of obstacle avoidance and route selection~\cite{fajen2003behavioral} have shown that the dynamics of perception and action consist of (i)~identifying the informational variables useful to guide behaviour and to regulate action, and (ii)~interacting with the environment using a particular set of dynamic behaviours. One possible policy descriptor allowing for this hierarchical control are \acfp{DMP}~\cite{ijspeert2013dynamical}. \Acp{DMP} are differential equations encoding kinematic control policies towards a goal attractor. Their transient behaviour can be shaped via a non-linear forcing term, which can be initialised via imitation learning and used to reproduce an observed motion while generalising to different start and goal locations, as well as task durations.

    A key feature of \acp{DMP} is that they allow for online modulation via coupling term functions that create a forcing term. Coupling terms have been exploited for many applications, such as avoidance of joint and workspace limits~\cite{gams2009line}, force control for environment interaction~\cite{gams2014coupling,sutanto2018learning}, dual-arm manipulation~\cite{gams2014coupling,pairet2018learning} and reactive obstacle avoidance~\cite{park2008movement,khansari2012dynamical,hoffmann2009biologically,rai2014learning,rai2017learning}. This work focuses on the latter challenge, which historically has been approached using potential fields~\cite{park2008movement,khansari2012dynamical}, analytical~\cite{hoffmann2009biologically} and learning methods~\cite{rai2014learning,rai2017learning} (see \sref{sec:background}). As further discussed in \sref{sec:background_discussion}, analytical formulations become less reactive for imminent collisions (dead-zone problem). Moreover, these approaches do not provide any guidance to the reactive behaviour, thus limiting their applicability to free-floating obstacles. Additionally, analytical formulations uniquely deal with point-mass obstacles and systems. In an attempt to address this latter issue, recent proposals learn coupling terms for a small set of obstacle geometries described by an array of markers on their surface~\cite{rai2014learning,rai2017learning}, but they fail to generalise actions to novel obstacles. These works are notable in learning the coupling terms from human demonstration. Nonetheless, providing a rich set of demonstrations involving various obstacles geometries can be time-consuming and prone to measurement noise.

    \begin{figure}[t!]
        \centering
        \includegraphics[width=8.5cm]{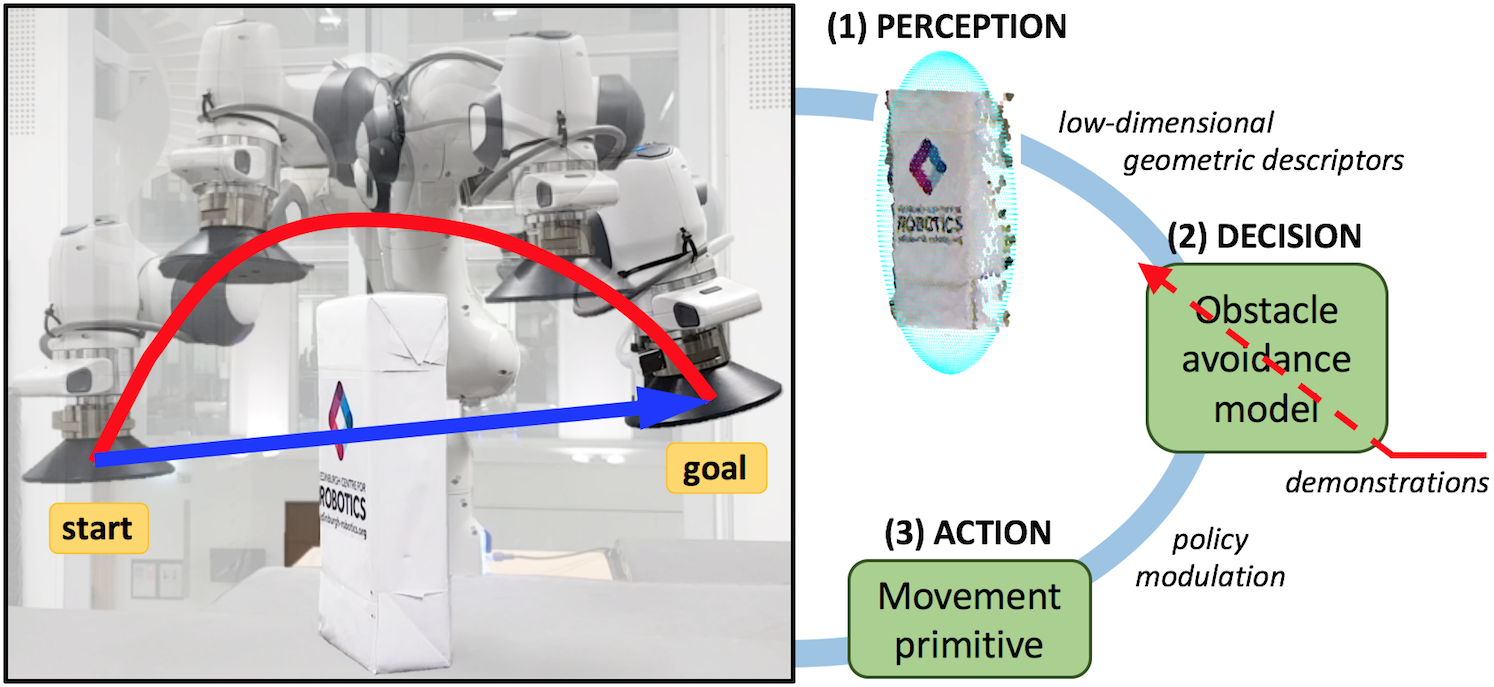}
        \caption{Proposed hierarchical framework for learning and producing generalisable obstacle avoidance behaviours. Pre-planned start-go-goal (blue) and modulated policy (red).}
        \label{fig:intro}
    \end{figure}

    
    This paper presents the hybrid \ac{DMP}-learning-based obstacle avoidance framework schematised in \fref{fig:intro}. The proposed approach addresses the limitations of the precedent works with a layered perception-decision-action analysis~\cite{fajen2003behavioral}. The main contributions at the action level (see \sref{sec:coupling_terms}) are (i)~reformulating the coupling terms to provide dead-zone free behaviours, and (ii)~guiding the obstacle avoidance reactivity to satisfy task-dependant constraints, while the main contributions at the perception-decision level (see \sref{sec:geometry}) are (iii)~regulating action according to the extracted unified system-obstacle low-dimensional geometric descriptor, and (iv)~learning to regulate the action level via exploration of the parameter space. The experimental evaluation reported in \sref{sec:results} demonstrates that the overall proposed approach generalises obstacle avoidance behaviours to novel scenarios, even when those involve multiple obstacles, or are uniquely described by partial visual-depth observations.
    

	\section{RELATED WORK} \label{sec:background}
    This paper proposes a reactive approach that endows a system with the ability to modulate its policy to avoid unexpected obstacles. The selected strategy uses \acp{DMP} for encoding any desired policy and defining an obstacle avoidance behaviour as a coupling term. This section introduces \acp{DMP} and coupling terms for obstacle avoidance as they constitute the fundamentals of this work.

    \subsection{Dynamic Movement Primitives} \label{sec:background_dmp}
        \Acp{DMP} are a versatile framework that encode primitive motions or policies as nonlinear functions called forcing terms~\cite{ijspeert2013dynamical}. The \acp{DMP} equations define the system's state transition, which can be converted into actuator commands by means of inverse kinematics and inverse dynamics. For a one-\ac{DoF} system, the system's state transition is described by the following set of nonlinear differential equations, known as the transformation system:
        \begin{align}
            \tau \dot{z} &= \alpha_x (\beta_x (g_{x} - \pos) - z) + \forcingterm + \couplingterm, \label{eq:dmp_1} \\
            \tau \vel &= z, \label{eq:dmp_2}
        \end{align}
        where $\tau$ is a scaling factor for time, $\pos$ is the system's position, $z$ and $\dot{z}$ respectively are the scaled velocity and acceleration, $\alpha_x$ and $\beta_x$ are constants defining the attraction dynamics towards the model's attractor $g_{x}$, and $\forcingterm$ and $\couplingterm$ are the forcing and coupling term, respectively.

        The forces generated by the forcing and coupling terms define the transient behaviour of the transformation system. It is common to model the forcing term $\forcingterm$ as a weighted linear combination of nonlinear \acp{RBF}. The evaluation of $\forcingterm$ at phase ${k \in \vk}$ is defined as:
        \begin{align}
            f(k) &= \frac{\sum_{i=1}^N w_i\Psi_i(k)}{\sum^N_{i=1} \Psi_i(k)} \; k, \label{eq:dmp_rbf}
        \end{align}
        \vspace{-0.4cm}
        \begin{align}
            \Psi_i(k) &= \exp\mathopen{}\left(-h_i(k-c_i)^2 \right)\mathclose{}, \label{eq:dmp_rbf_def}
        \end{align}
        where $c_i$ and $h_i > 0$ are the centres and widths, respectively, of the $i \in [1, \; N]$ \acp{RBF}, which are weighted by $w_i$ and distributed along the trajectory. The weights can be initialised via imitation learning and used to reproduce the motion with some generalisation capabilities to changes in start and goal positions. The duration of the motion can be adjusted by the scaling factor $\tau$, which modifies the canonical system defining the transient behaviour of the phase variable $k$ as:
        \begin{align}
            \tau \dot{k} = -\alpha_k k,
        \label{eq:canonicalsystem}
        \end{align}
        where the initial value of the motion's phase $\vk(0) = 1$ and $\alpha_k$ is a positive constant.

        A common strategy to extend the spatial generalisation capabilities of \acp{DMP} is to reference them in a local frame, whose pose in the space is task-dependent~\cite{ijspeert2013dynamical,rai2017learning}. In this work's context, the unit vectors of the local frame are defined as follows: the x-axis points from the start position towards the goal position, the z-axis points upwards and is orthogonal to the local x-axis, and the y-axis is orthogonal to both local x-axis and z-axis following the right-hand convention.

        A robot with multiple \acp{DoF} uses a transformation system for each \ac{DoF}, but they all share the same canonical system.

    \subsection{Coupling Terms for Obstacle Avoidance}
        \begin{figure}[b!]
			\vspace{-0.2cm}
            \centering
            \subfloat[]{\includegraphics[clip,trim={2.65cm 0.6cm 0.8cm 0},width=4.5cm]{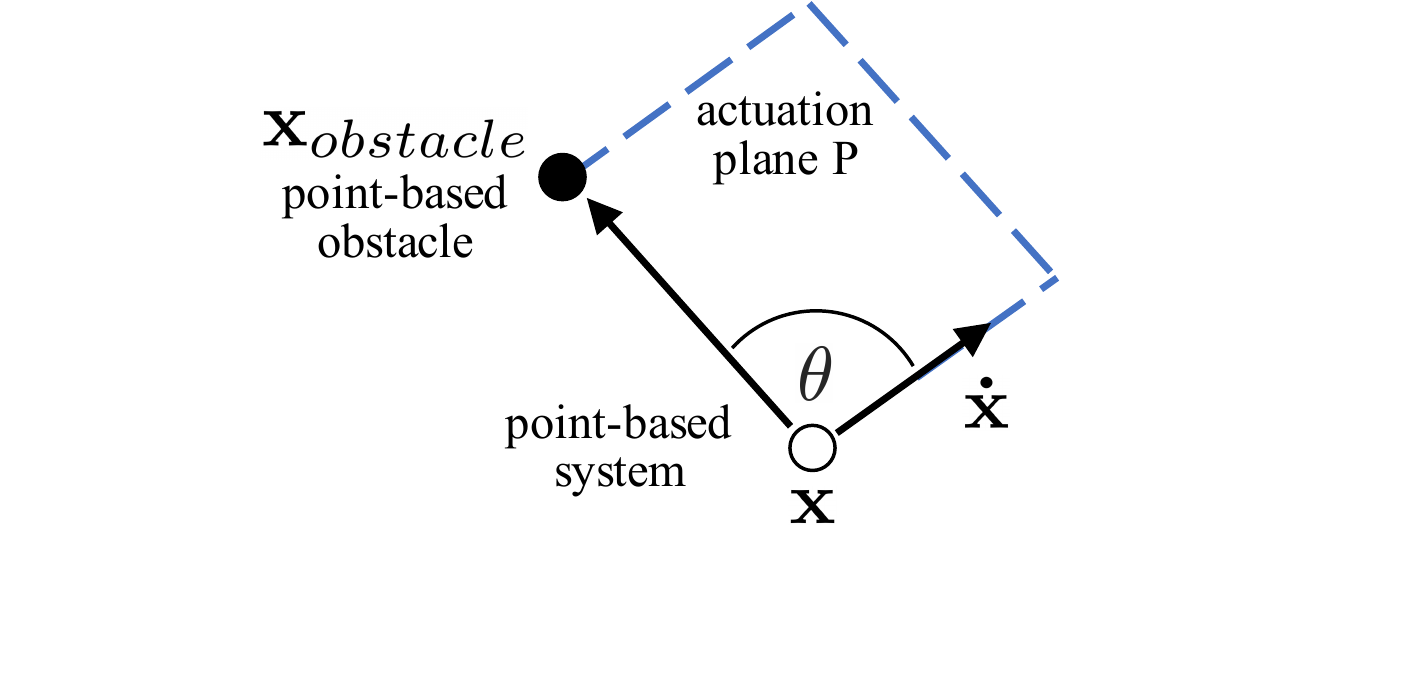} \label{fig:oa_scheme}}
            \hspace{-1cm}
            \subfloat[]{\includegraphics[height=2.8cm]{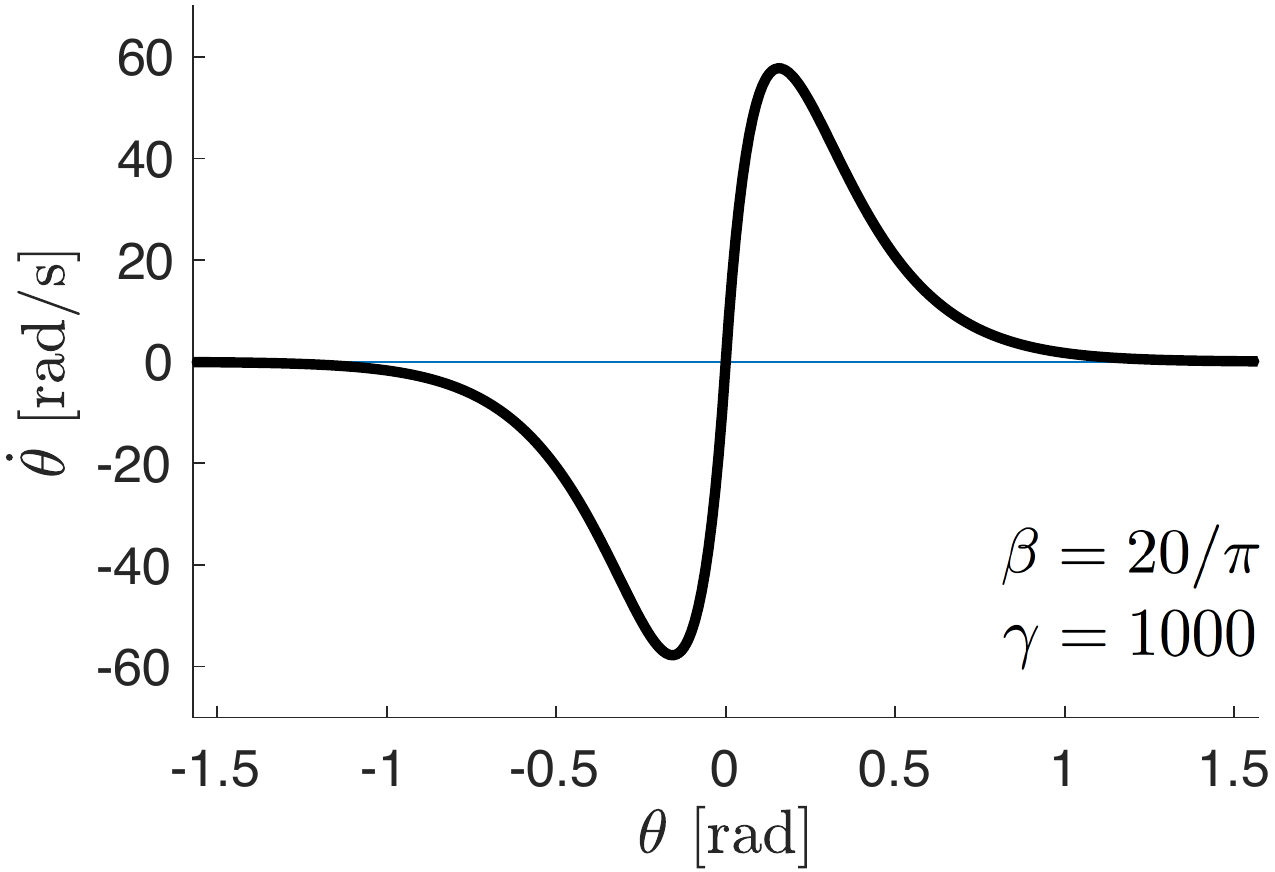} \label{fig:oa_original}}
            \caption{Original coupling terms for obstacle avoidance~\cite{hoffmann2009biologically}. (a)~Heading angle $\theta$ according to velocity vector $\vvel$ and the relative obstacle-system position in $\pplane$-plane. (b)~Change of steering angle $\dot{\theta}$ subject to heading angle $\theta$ as defined by~\eref{eq:oa_original}.}
            \label{fig:oa_overall_original}
        \end{figure}


        Early coupling terms for obstacle avoidance were formulated as repulsive potential fields~\cite{park2008movement,khansari2012dynamical}. Potential fields suffer from local minima and can be computationally expensive to calculate on the fly. Alternatively, some coupling terms analytically formalise the influence of an obstacle on the system's behaviour~\cite{hoffmann2009biologically}. As depicted in \fref{fig:oa_scheme}, a point-mass system with position~${\vpos\in\real^3}$ and velocity~${\vvel \in \real^3}$ has a heading~${\theta \in \rotmat(2)}$ towards a point-mass obstacle. To avoid a collision, the coupling term generates a repulsive force:
        \begin{align}
             \couplingterm = \mathbf{R} \; \vvel \; \dot{\theta}, \label{eq:oa_main}
        \end{align}
        where ${\mathbf{R} \in \rotmat(3)}$ is a $\pi/2$ rotation matrix around the vector ${\mathbf{r} = (\vpos_{obstacle} - \vpos) \times \vvel}$. The respective obstacle-system position ${\vpos_{obstacle} - \vpos}$ and the system's velocity $\vvel$ define the plane $\pplane \in \mathbb{R}^2$ where the system is desired to steer away from the obstacle with a turning velocity $\dot{\theta}$ defined as:
        \begin{align}
            \dot{\theta} = \gamma \; \theta \exp \!\left(-\beta \; |\theta| \right)\!, \label{eq:oa_original}
        \end{align}
        where $\gamma$ and $\beta$ respectively scale and shape the mapping ${\theta \rightarrow \dot{\theta}}$ defined in \eref{eq:oa_original} and represented in \fref{fig:oa_original}.


        Building on \eref{eq:oa_main}-\eref{eq:oa_original}, human demonstrations were used to retrieve the required parameters to circumvent two non-point obstacles, particularly a sphere and a cylinder~\cite{rai2014learning}. More recently, coupling terms were formulated as independent \acp{NN} modelling the desired obstacle avoidance behaviour for a sphere, a cylinder and a cube~\cite{rai2017learning}. These methods do not provide any strategy to avoid obstacles not observed in training time, and they rely on markers identifying an obstacle's boundaries. Their evaluations are conducted either in simulation or in single-obstacle scenarios. Hence, their performance in realistic scenarios is yet to be tested.

    \subsection{Discussion and Contribution} \label{sec:background_discussion}

        State-of-the-art on coupling terms modelling obstacle avoidance behaviours suffers from four major limitations. First, as illustrated in \fref{fig:oa}, the analytical term \eref{eq:oa_main}-\eref{eq:oa_original} has a dead-zone where the system becomes less reactive as the heading towards the obstacle narrows, thus compromising the method's reliability. Second, there is no strategy to guide the behaviour's reactivity towards a preferred route to circumnavigate an obstacle. For example, in the scenario depicted in \fref{fig:intro}, there is no constraint on the reactive behaviour preventing the system from hitting the table. Third, when attempting to deal with non-point obstacles, their performance drastically decreases for novel scenarios due to the absence of global features identifying the obstacle geometry during the learning process. Fourth, these works learn the coupling terms from demonstration, which can be time-consuming and prone to measurement noise.

        All these issues are jointly addressed within the proposed hierarchical framework, which hybridises the versatility of \acp{DMP} and the strengths of learning techniques. Specifically, in \sref{sec:coupling_terms}, \eref{eq:oa_main}-\eref{eq:oa_original} is reformulated at the action level as a conjunction of coupling terms whose obstacle avoidance behaviour is dead-zone free and can be guided. Then, in \sref{sec:geometry}, the formalised action level is exploited to learn via exploration of the parameter space how to regulate the behaviour subject to both the end-effector's and obstacle's geometric properties. This work considers a unified system-obstacle low-dimensional geometric descriptors identifying the relevant features to the action level, thus allowing for enhanced generalisation even in novel real-world scenarios.

	\section{COUPLING TERMS for DEAD-ZONE FREE and GUIDED OBSTACLE AVOIDANCE} \label{sec:coupling_terms}
    The proposed hierarchical framework to learn and produce generalisable obstacle avoidance behaviours regardless of the scenario comprises three layers. The \ac{DMP}-based action level is formalised as a composition of two coupling terms which (i)~generate robust obstacle avoidance behaviours, and (ii)~guide these in a particular direction of the task space. The parametrisation needs of these terms allow for regulating their actuation scope via reasoning at the decision level.

    \subsection{Inherently Robust Obstacle Avoidance} \label{sec:ct_avoiding}
        Current coupling terms for obstacle avoidance in the literature suffer from dead-zones, i.e. a heading range towards the obstacle for which the system becomes incoherently less reactive. Ideally, the expected behaviour of those terms would be to become more reactive as (i)~the heading of the system is more aligned towards an obstacle, and (ii)~the system-obstacle distance is smaller. Bearing these conditions in mind, the coupling term in \eref{eq:oa_main}-\eref{eq:oa_original} is reformulated as:
        \begin{align}
            \couplingtermOA = \mathbf{R} \; \vvel \; \newgamma\sign(\theta)\exp\!\left(\!\!-\frac{\theta^2}{\newbeta^2} \!\right)\exp \!\left(\!-\newkappa \thinspace d^2 \right)\!, \label{eq:oa_complete}
        \end{align}
        where ${\newgamma\sign(\theta)\exp\!\left(\!-{\theta^2}/{\newbeta^2}\right)\!}$ addresses the first issue by shaping the absolute change of steering angle as a zero-mean Gaussian-bell function, and ${\exp\!\left(\!-\newkappa \, d^2 \right)}\!$ tackles the second requirement by regulating the coupling term effect according to a parameter~$k$ and the system-obstacle distance~$d$.

\begin{figure}[t]
            \centering
            \subfloat[]{\includegraphics[height=1.66cm]{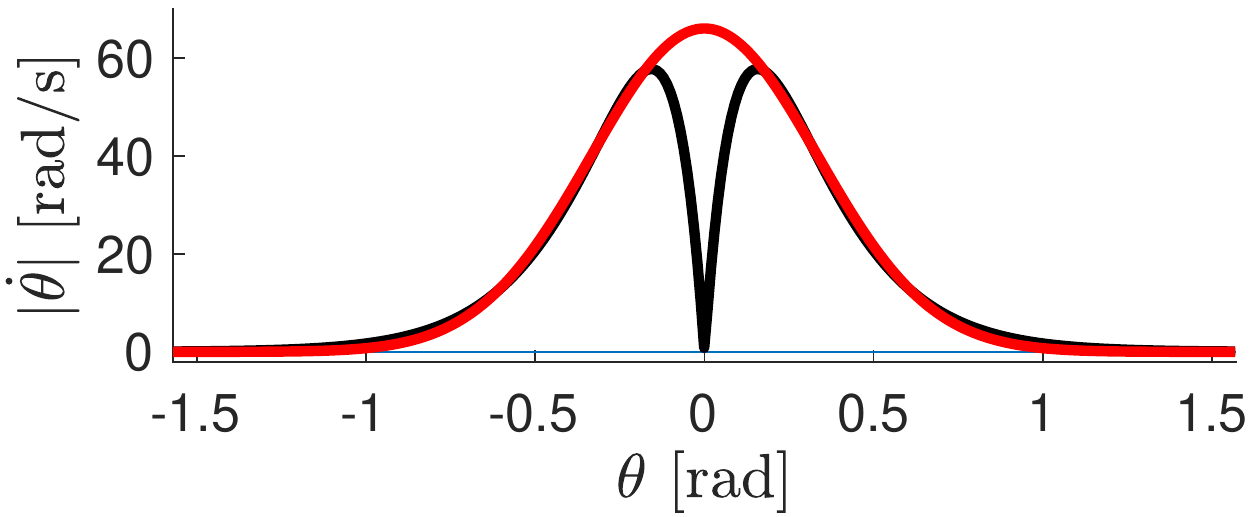} \label{fig:oa_comparison}}
            \subfloat[]{\includegraphics[height=1.66cm]{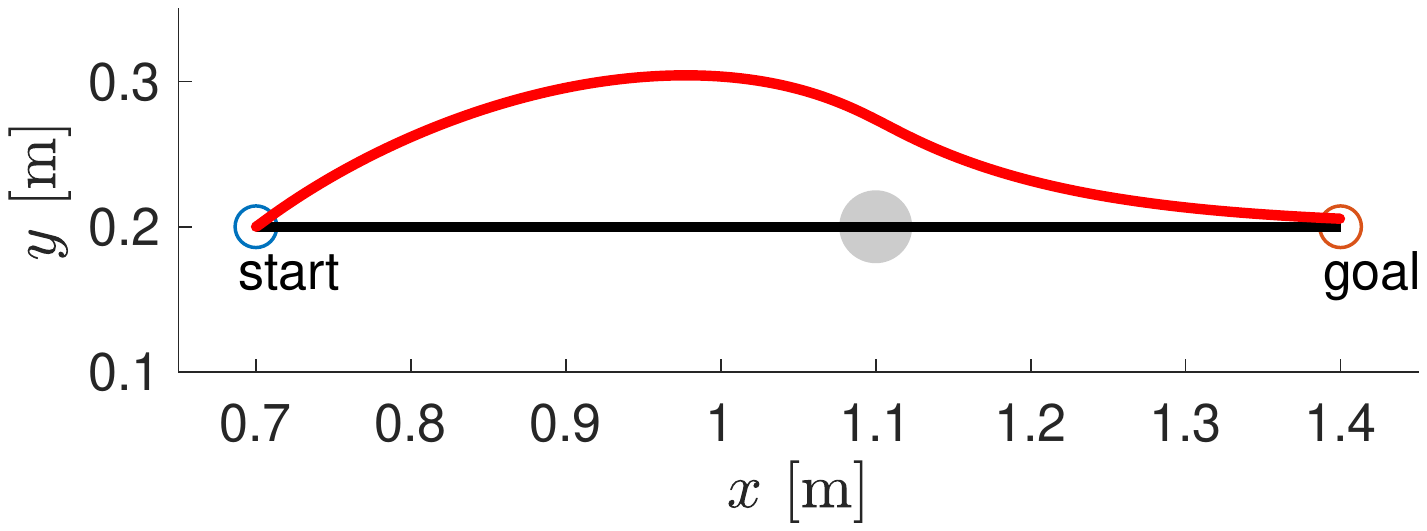} \label{fig:oa_example}}
            \caption{Dead-zone issue in the original~\eref{eq:oa_main}-\eref{eq:oa_original} (black) and proposed~\eref{eq:oa_complete} (red) coupling terms. (a)~\eref{eq:oa_complete} reacts for narrow headings towards the obstacle. (b)~\eref{eq:oa_main}-\eref{eq:oa_original} fails where~\eref{eq:oa_complete} smoothly circumvents the point-mass obstacle (grey circle).}
            \label{fig:oa}
        \end{figure}

        \fref{fig:oa} highlights the increase in robustness of the formulated coupling term~\eref{eq:oa_complete} in contrast to the original term~\eref{eq:oa_main}-\eref{eq:oa_original}. While the original coupling term (black curves) produces low reactivity for narrow headings towards an obstacle, the dead-zone free proposal (red curves) reacts the most (see \fref{fig:oa_comparison}).
        This reformulation has a significant impact in the task space, where~\eref{eq:oa_complete} succeeds on a scenario where~\eref{eq:oa_main}-\eref{eq:oa_original} fails to generate an obstacle avoidance behaviour which does not collide with the point-mass obstacle (see \fref{fig:oa_example}).

    \subsection{Guiding the Obstacle Avoidance Reactivity} \label{sec:ct_guiding}
        \begin{figure}[b!]
			\vspace{-0.2cm}
            \centering
            \subfloat[]{\includegraphics[height=2cm]{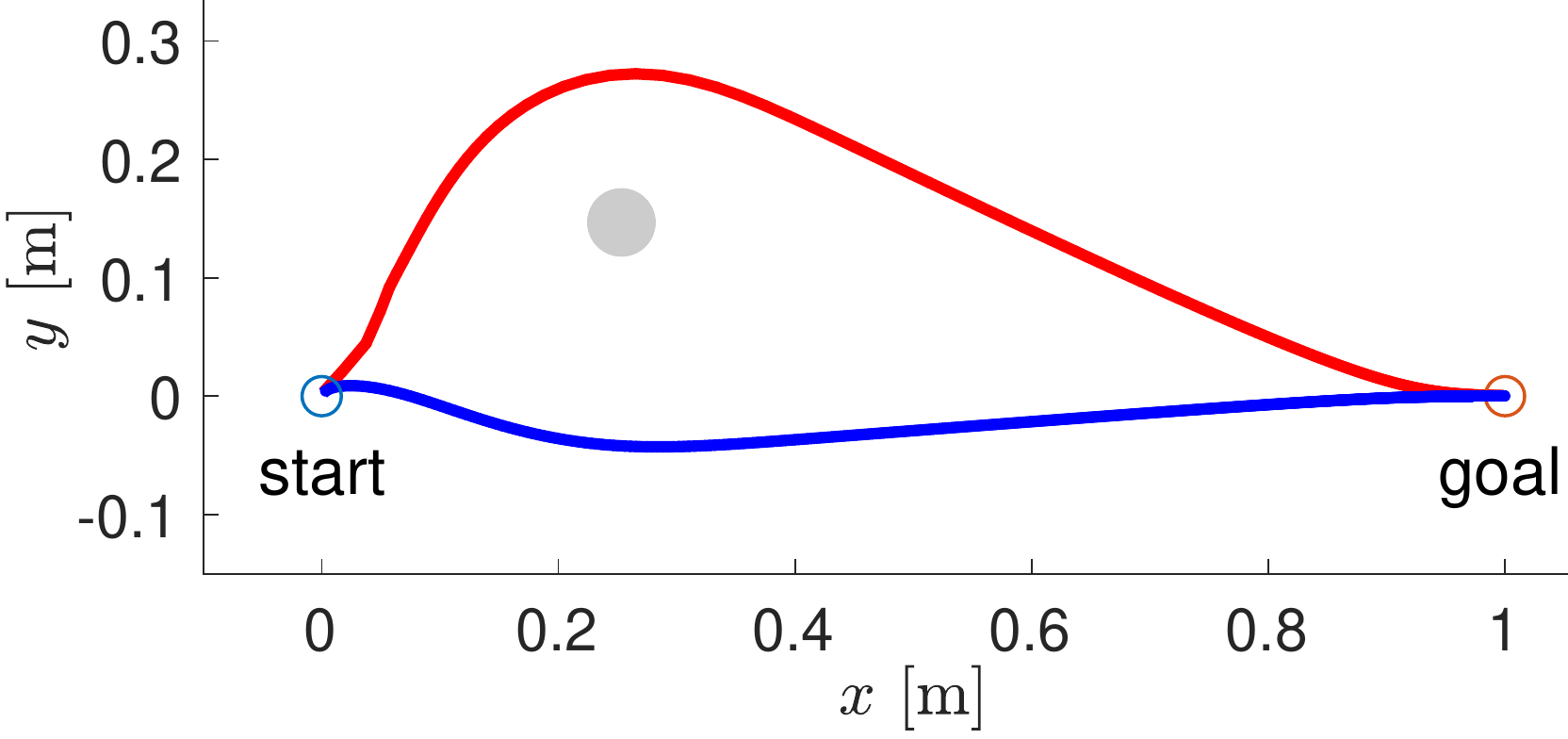} \label{fig:rs_one_obs}}
            \subfloat[]{\includegraphics[height=2cm]{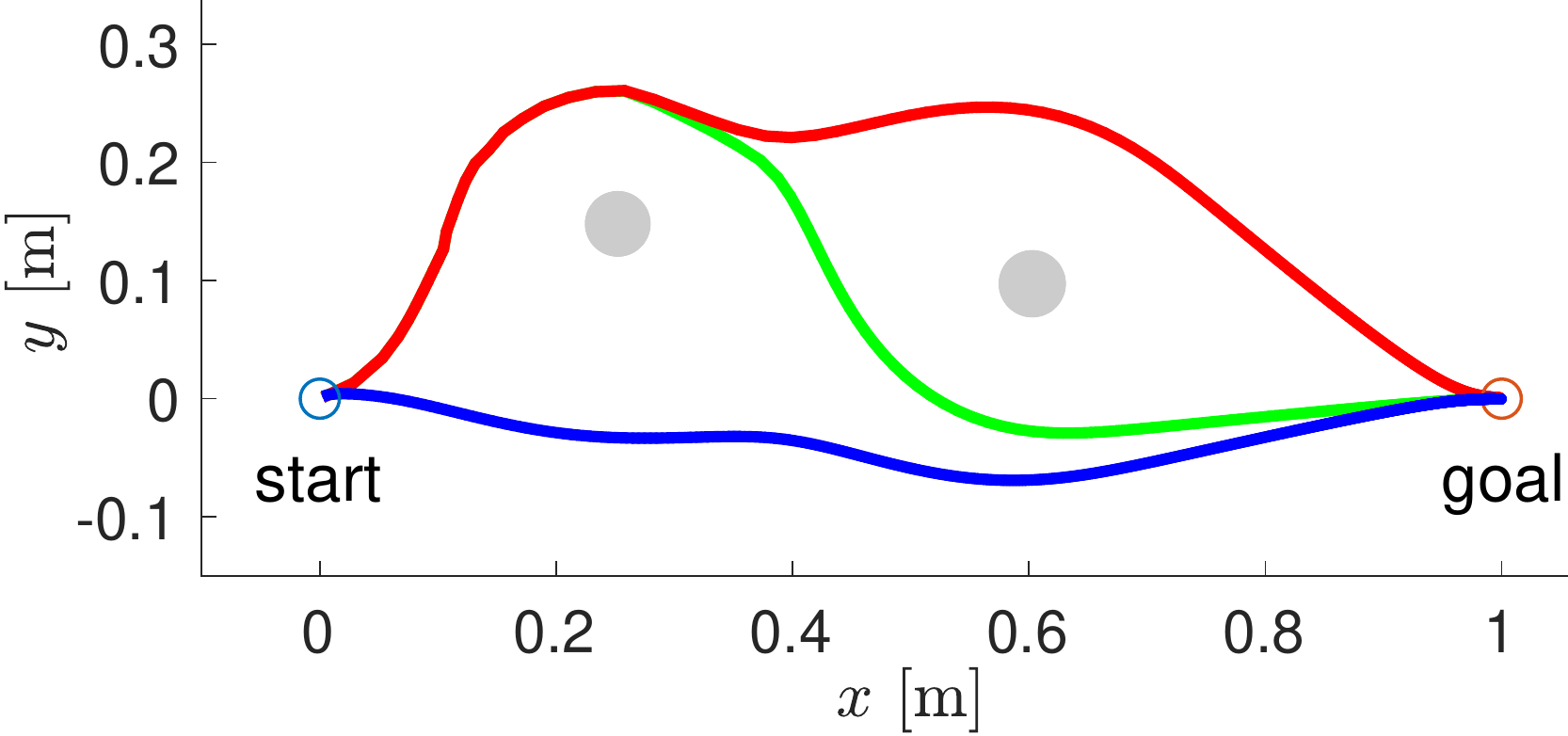} \label{fig:rs_two_obs}}
            \caption{Route selection for obstacle avoidance in (a)~single and (b)~multi-object setups. Reactive avoidance (blue), one-time decision (red), two-times decision (green).}
            \label{fig:direction_guidance}
        \end{figure}

        The velocity vector $\vvel$ of a point-mass system also represents the system's orientation. Consequently, $\vvel$ plays a critical role in determining both the actuation $\pplane$-plane and the direction of turning~$\dot{\theta}$. Overall, the behaviour encapsulated in~\eref{eq:oa_complete} consists of turning to the opposite direction where the obstacle is with respect to the system's heading or velocity vector $\vvel$. Although this reactive motion might be the safest behaviour in front of an obstacle, there are many situations where guiding the system towards a particular route might be of interest, such as in constrained environments or when aiming for a trajectory providing a minimum cost. 

        Given the influence of the system's heading $\vvel$ on the overall obstacle avoidance reaction, it is natural to modulate $\vvel$ to guide the reactivity of~\eref{eq:oa_complete} through a preferred route. Within the \ac{DMP} motion descriptor, this can be formulated through a coupling term that creates an attractive forcing term to reduce the heading error $\hat{\theta}$ between the current $\vvel$ and a desired $\vvel_d$ system's direction as:
        \begin{align}
             \couplingtermHG = \mathbf{R}^\prime \; \vvel \; \newgamma \; \hat{\theta} \; \exp \! \left(1 + \newkappa \thinspace d^2 \right) \label{eq:oa_rs}
        \end{align}
        where ${\mathbf{R}^\prime \in \rotmat(3)}$ is a $\pi/2$ rotation matrix around the vector ${\mathbf{r}^\prime = \vvel \times \vvel_d}$, and the term ${\newgamma\exp\!\left(1 + \newkappa \thinspace d^2 \right)}$ ensures that \eref{eq:oa_complete} and \eref{eq:oa_rs} act in counterphase when parameterised for the same $\newgamma$~and~$\newkappa$. This is, \eref{eq:oa_rs} uniquely modifies the system's heading when not in proximity to obstacles, where \eref{eq:oa_complete} takes over the control to ensure the system's safety.

    \subsection{Coupling Terms Composition} \label{sec:ct_composition}
        \fref{fig:direction_guidance} depicts the significance of using~\eref{eq:oa_complete} in conjunction with~\eref{eq:oa_rs} to perform route selection of obstacle avoidance. This is formalised within the \ac{DMP} in \eref{eq:dmp_1}-\eref{eq:dmp_2} as the composition of coupling terms ${\couplingterm = \sum_i \couplingtermOAi + \couplingtermHGi}$, where $\couplingtermOAi$ and $\couplingtermHGi$ generate the corresponding forcing terms with respect to the $i^{th}$ obstacle in the scenario. This composition allows, in a single-obstacle scenario (see \fref{fig:rs_one_obs}), to guide the reactive behaviour (blue trajectory) in a different direction (red trajectory) by temporarily defining the initial desired heading towards the upper part of the task space. The same applies to multi-obstacle environments (see \fref{fig:rs_two_obs}), where the system's heading can be modified at multiple decision points to obtain a preferred route (green trajectory). In both scenarios, the actuation scope of the coupling term for guiding the system was set manually for illustration purposes. Alternatively, these decision points could be defined by a task-dependant module.

    \subsection{Proof of Lyapunov's Stability} \label{sec:ct_proof}
        The addition of coupling terms can imperil the inherent stability properties of \acp{DMP}~\cite{ijspeert2013dynamical}. Authors in~\cite{rai2014learning} proved with Lyapunov's theory that the overall dynamical system remains stable when the coupling terms generate a forcing term orthogonal to the system's velocity vector. The coupling terms formulated in \eref{eq:oa_complete} and \eref{eq:oa_rs} satisfy this condition, therefore proving the global stability of the proposed action level.


\section{LEARNING OBSTACLE AVOIDANCE \\ for NON-POINT OBJECTS} \label{sec:geometry}
    The set of coupling terms formalised in the previous section efficiently generates guided collision-free trajectories for point-mass objects, i.e. obstacles and systems. Nonetheless, objects in real-world scenarios present different shapes and sizes. This section details the encoding of objects as low-dimensional geometric descriptors, which allows for (i)~the design of a learning module that regulates the action level to generalise over different obstacle geometries while considering the system's geometry, and (ii)~the use of heuristics to rapidly perform route selection in constrained environments.


    \subsection{Superquadrics as Geometric Approximates} \label{sec:geometry_sq}

        Objects obstructing the execution of a policy might present different shapes and dimensions. This geometric diversity complicates the design of an intelligent module able to generalise obstacle avoidance behaviours across geometries~\cite{rai2017learning}. This work considers global features to approximate the geometric properties of an object. One possible encoding strategy are superquadrics~\cite{barr1981superquadrics}, which have been used, among others,
        to ease the computation of system-obstacle distances~\cite{perdereau2002real}, and to generate repulsive potential fields~\cite{khatib1986real}. Alternatively to these task space applications, this work is interested in the low-dimensional parametric encoding of such geometric approximate, which is defined as:
        \begin{align}
            \!\!\!\!\!F(x,y,z,\boldsymbol{\lambda})\!: \left( \left( \frac{x}{\lambda_1} \right)^{\frac{2}{\lambda_5}} \!\! + \left( \frac{y}{\lambda_2} \right)^{\frac{2}{\lambda_5}}\right)^{\frac{\lambda_5}{\lambda_4}} \!\! + \left( \frac{z}{\lambda_3} \right)^{\frac{2}{\lambda_4}}\!\!\!,
            \label{eq:sq}
        \end{align}
        where $F(\boldsymbol{\cdot})$ defines whether a given 3D point ${(x, y, z)}$ lies inside (${F<1}$), outside (${F>1}$), or on the surface (${F=1}$) of a superquadric described by ${\boldsymbol{\lambda} = [\lambda_1, ..., \lambda_5]}$. In particular, ${(\lambda_1, \lambda_2, \lambda_3)}$ set the superquadric semi-axes lengths, and ${(\lambda_4, \lambda_5)}$ parameters define the superquadric shape.

        The parameter vector ${\boldsymbol{\lambda}}$ can be estimated from a discrete representation of the obstacle's surface by minimisation of:
        \begin{align}
            \min_{\boldsymbol{\lambda}} \sum_{i=1}^{N} \left(\sqrt{\lambda_1 \lambda_2 \lambda_3} \left(F(x_i,y_i,z_i,\boldsymbol{\lambda}) - 1 \right) \right)\!, \label{eq:sq_min}
        \end{align}
        where ${\sqrt{\lambda_1 \lambda_2 \lambda_3}}$ penalises the fitting of large superquadrics.

    \subsection{Unified Low-dimensional Geometric Descriptors} \label{sec:geometry_unified}


        \begin{figure}[b!]
			\vspace{-0.2cm}
            \centering
            \includegraphics[width=8.5cm]{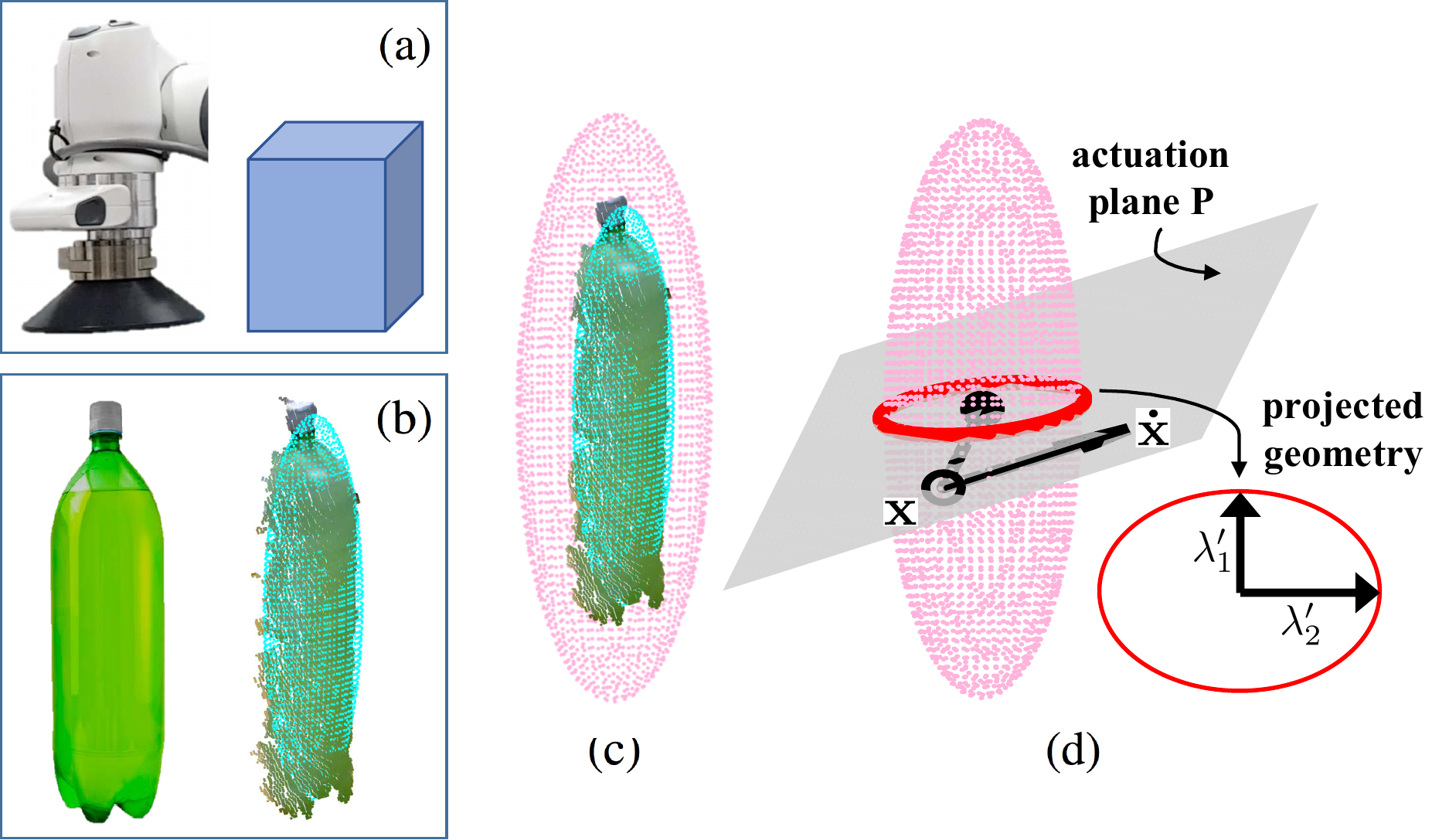}
            \subfloat{\label{fig:ga_complete_a}}
            \subfloat{\label{fig:ga_complete_b}}
            \subfloat{\label{fig:ga_complete_c}}
            \subfloat{\label{fig:ga_complete_d}}
            \caption{Extraction of unified low-dimensional descriptors accounting for the (a)~end-effector's and (b)~obstacle's geometry. (c)~Ellipsoid (rose) fitting the dilated obstacle cloud. (d)~Relevant descriptor $\vprojectedgeometry$ along the $\pplane$-plane (red ellipse).}
            \label{fig:ga_complete}
        \end{figure}

        The process in \eref{eq:sq}-\eref{eq:sq_min} provides a geometrical descriptor~${\boldsymbol{\lambda}}$ from a discrete representation of an object. However, it is of interest to obtain a descriptor accounting for both the system's and obstacle's geometry. \fref{fig:ga_complete} schematises the extraction of a unified obstacle-system low-dimensional geometric descriptor. An approximate of the system's geometry (see blue prism in \fref{fig:ga_complete_a}) is used to dilate~\cite{lozano1990spatial,huber2019avoidance} the obstacle's discrete representation (see \fref{fig:ga_complete_b}). The dilated obstacle representation is then encoded using \eref{eq:sq_min} while imposing ${\lambda_4 = \lambda_5 = 1}$, i.e. restricting the superquadric to shape as an ellipsoid. \fref{fig:ga_complete_c} portrays the significance on the descriptor's difference when considering the raw obstacle representation (blue ellipsoid) and its dilated version (rose ellipsoid). Interestingly, ellipsoids hold the property that any random projection or section of these results in an ellipse, providing a strategy to extract the unified obstacle-system's geometric features relevant to the obstacle avoidance coupling term. This is, the $\pplane$-plane defined by the respective obstacle-system position ${\vpos_{obstacle} - \vpos}$ and the system's heading $\vvel$, intersects the unified geometric approximation. Thus, the descriptor ${\boldsymbol{\lambda}}$ can be further reduced to ${\vprojectedgeometry =(\projectedwidth, \projectedheight)\in \mathbb{R}^2}$ such that ${\vprojectedgeometry = g(\boldsymbol{\lambda})}$ where ${g(\boldsymbol{\cdot})\!:\mathbb{R}^5\rightarrow\mathbb{R}^2}$ maps an arbitrary vector onto the $\pplane$-plane. The resulting low-dimensional descriptor is an ellipse laying on the $\pplane$-plane with semi-axis lengths $\vprojectedgeometry$ (see \fref{fig:ga_complete_d}).

    \subsection{Geometry-conditioned Parameter Regressor} \label{sec:geometry_rc}
        Leveraging the unified low-dimensional descriptor $\vprojectedgeometry$ from \sref{sec:geometry_unified}, this section proposes a method to learn the correspondence between $\vprojectedgeometry$ and the non-independent parameters ${(\newgamma, \newbeta, \newkappa)}$ of the coupling term, subject to a user-defined clearance~$\clearance$, i.e. the minimum distance between the end-effector and the obstacle. This \acl{MTR} problem is formulated as a \ac{RC}~\cite{spyromitros2016multi}, which defines an ordered chain ${U=(Y_1, Y_2, Y_3)}$ of single target regressions. This is, given an input vector ${\inputvector=\{\vprojectedgeometry,\clearance\}}$, the proposed \ac{RC}-based learning module is composed of three models: ${Y_1\!: \inputvector \rightarrow \newkappa}$ adjusts the actuation span of the coupling term, ${Y_2\!: (\inputvector, \newkappa) \rightarrow \newbeta}$ regulates the relevance of the relative system-obstacle heading, and finally ${Y_3\!: (\inputvector, \newkappa, \newbeta) \rightarrow \newgamma}$ tunes the strength of the behaviour. Each regressor $Y_i$ is modelled as a \ac{NN} which provides a powerful strategy to learn and represent approximations to non-linear mappings, and is suitable for reactive decisions due to its rapid response. Considering the relevance of the input features, each \ac{NN} regressor is arranged with four layers; the hidden layers are hyperbolic tangent sigmoid units, and the output layer is a log-sigmoid to avoid negative settings of the targets.

        It should be noted that the regulation of the action level formalised in \sref{sec:coupling_terms} is conducted along the $\pplane$-plane. As explained previously in \sref{sec:geometry_sq}, this sub-space contains all essential information to circumnavigate an obstacle and is efficiently defined using the relative system-obstacle state. Namely, changes in the obstacle avoidance scene such as different start and goal positions, obstacle location and geometries do not alter the encoding of the problem in the $\pplane$-plane. Therefore, the prediction capabilities of the designed \ac{RC}-based learning module extend to a wide range of setups, including in the presence of multiple obstacles in the scene.

    \subsection{Route Selection via Heuristic Cost Rings} \label{sec:geometry_route_selection}

        Real-world environments and physical systems constrain the amount of feasible reactive behaviours. Exhaustively evaluating all possible directions in $\rotmat(3)$ which satisfy these additional constraints can slow the decision response. To ease the reasoning complexity of the route selection problem, this work proposes a twofold heuristic analysis called cost rings which (i)~considers an orthographic projection of the obstacle onto the ${\yzplane\text{-plane} \in \real^2}$ of the local frame, i.e. confining the direction space ${\boldsymbol{\omega}\in\rotmat(2)}$, to then efficiently (ii)~find the obstacle avoidance direction ${\omega_d}$ minimising a metric ${\cost(\boldsymbol{\omega})}$.
        The resulting direction ${\omega_d}$ is used with the coupling terms composition formulated in~\sref{sec:ct_composition} to guide the obstacle avoidance behaviour towards ${\omega_d}$.

\begin{figure}[t!]
            \centering
            \subfloat[]{\includegraphics[clip,trim={0cm 0cm 0cm 0},width=3.8cm]{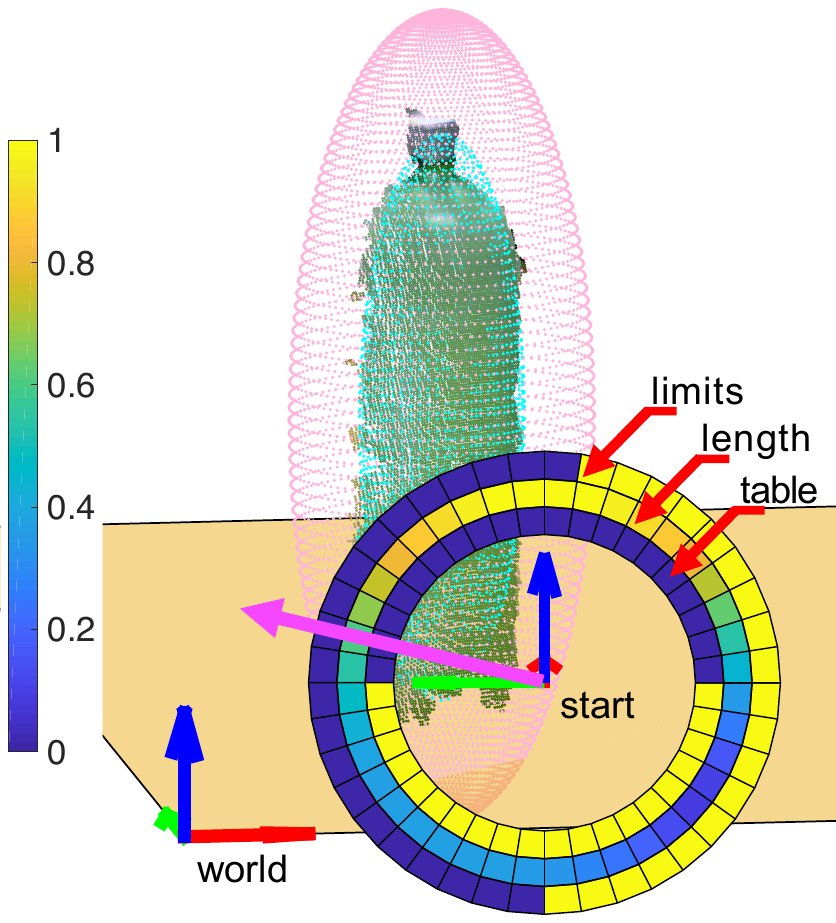} \label{fig:cost_rings_heuristics}}
            \quad
            \subfloat[]{\includegraphics[clip,trim={0cm -0.15cm 0cm 0},width=4cm]{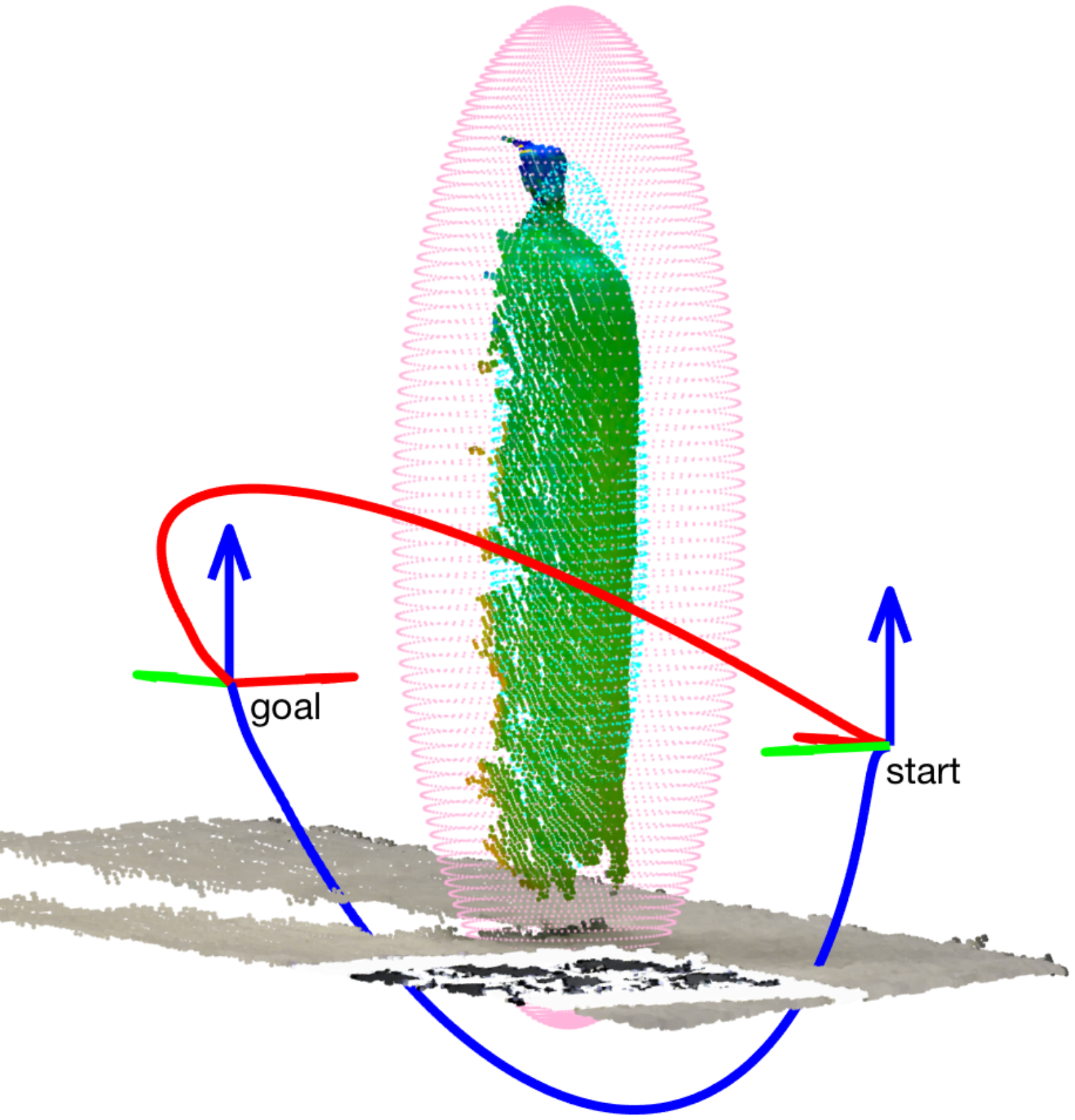} \label{fig:cost_rings_workspace}}
            \caption{Route selection via heuristic rings. (a)~Cost evaluation on the local $\yzplane$-plane to penalise workspace limits, long trajectories, and collisions with the table. Overall best direction is marked in magenta. (b)~A reactive behaviour (blue) would lead the system colliding with the table, whereas the guided behaviour (red) generates the route with lowest cost.}
            \label{fig:cost_rings}
			\vspace{-0.3cm}
        \end{figure}

        The advantage of route selection via heuristic cost rings is exemplified in \fref{fig:cost_rings}, where the path cost ${\cost(\boldsymbol{\omega})}$ is determined according to three metrics: (i)~the physical constraints imposed by the table ${\cost_{table}(\boldsymbol{\omega})}$, (ii)~the length of the trajectory ${\cost_{length}(\boldsymbol{\omega})}$, and (iii)~the robot's workspace limit ${\cost_{limits}(\boldsymbol{\omega})}$, such that $\omega_d$ can be found by minimisation of:
        \begin{equation}
            \min_{\boldsymbol{\omega}} \cost_{table}(\boldsymbol{\omega}) + \cost_{length}(\boldsymbol{\omega}) + \cost_{limits}(\boldsymbol{\omega}),
        \end{equation}
        where ${\cost_{table}(\boldsymbol{\omega}) = 1}$ if the end-effector would collide with the table and $0$ otherwise, ${\cost_{length}(\boldsymbol{\omega}) \in [0,1]}$ is the normalised trajectory length, and ${\cost_{limits}(\boldsymbol{\omega})=1}$ if the end-effector would move outside of its workspace and $0$ otherwise. \fref{fig:cost_rings_heuristics} illustrates these estimated costs rings and the resulting direction ${\omega_d \in \boldsymbol{\omega}}$ (magenta) with minimum cost. As depicted in \fref{fig:cost_rings_workspace}, using this reasoning to initially guide the behaviour enables the system to avoid the obstacle in the direction with lowest cost (red), whereas the non-guided reactive behaviour leads with collision with the table (blue).

    \subsection{Convergence to Goal} \label{sec:geometry_route_goal_convergence}
        The required path $\hat{\pi}$ to avoid obstacles may be longer than the pre-planned trajectory ${\pi}$, thus needing more time to finalise the encoded task. This fact is especially critical when dealing with non-point objects as failing to account for this can imperil convergence to the desired goal~\cite{rai2017learning}. To address this issue, this work regulates the \ac{DMP} duration by scaling ${\tau = length(\mathbf{\hat{\pi}}) / length(\mathbf{\pi})}$, i.e. an approximate of the increase of trajectory length. Here, ${length(\mathbf{\hat{\pi}})}$ is estimated with linear interpolation of the finite sequence of $\real^3$ points ${\{\vpos_s, \vpos_p^1, ..., \vpos_p^N, \mathbf{g}_{x}\}}$, where ${\vpos_s}$ and ${\mathbf{g}_{x}}$ are the start and goal positions, and ${\mathbf{x}_p^i}$ is the extreme point of the ellipse encoding the
        ${i \in [1, \; N]}$ dilated obstacle's geometry along its $\pplane$-plane.

    \section{EXPERIMENTAL EVALUATION} \label{sec:results}
    The proposed framework has been evaluated in simulated environments and on a physical system. This section first explains the training of the \ac{RC} model via exploration of the parameter space. Thereafter, it reports the performance and generalisation capabilities of the proposed approach in familiar and novel obstacle avoidance settings. Finally, this section details the deployment of the proposed framework on an anthropomorphic Franka Emika Panda arm engaged in a start-to-goal policy in the presence of unplanned obstacles.

	An extended illustration of the experimental evaluation is documented in: \mbox{\url{https://youtu.be/lym5cCbjI3k}},
	and the corresponding source code can be found in: \mbox{\url{https://github.com/ericpairet/ral_2019}}.

    \subsection{Training the \ac{RC}-based Learning Module} \label{sec:results_training}
        \begin{table}[b]
			\vspace{-0.2cm}
            \centering
            \begin{tabular}{lcccccc}
                \toprule
                & \multicolumn{2}{c}{$\text{NMSE}(Y_1)$} & \multicolumn{2}{c}{$\text{NMSE}(Y_2)$} & \multicolumn{2}{c}{$\text{NMSE}(Y_3)$} \\
                \cmidrule(lr){2-3} \cmidrule(lr){4-5} \cmidrule(lr){6-7}
                & train & test & train & test & train & test \\
                \cmidrule{2-7}
                \ac{RC}($\vprojectedgeometry$) & 0.539 & 0.543 & 0.802 & 0.802 & 0.893 & 0.897 \\
                \ac{RC}($\vprojectedgeometry$, $\clearance$) & 0.251 & 0.253 & 0.244 & 0.243 & 1.6e-4 & 1.7e-4 \\ 
                \bottomrule
            \end{tabular}
            \caption{Prediction error on every single target of the two modelled \ac{RC} architectures for the training and test datasets.}
            \label{tab:nn_training}
        \end{table}

        This work has designed a \ac{RC}-based learning module to regulate the action level according to a unified obstacle-system descriptor $\vprojectedgeometry$ and a possible clearance constraint $\clearance$. The
        unconstrained model is denoted as \ac{RC}($\vprojectedgeometry$), while the constrained model is referred to as \ac{RC}($\vprojectedgeometry,\clearance$). The training of these models is conducted leveraging the knowledge of the action level to create a synthetic dataset via exploration of the parameter space. This is, given different obstacle avoidance scenarios, training explores the parameters ${\{\newgamma, \newbeta, \newkappa\}}$ of the coupling term~\eref{eq:oa_complete} generating a collision-free trajectory.


        Bearing in mind that the learning module uniquely regulates the action level along its plane of actuation, $100$ synthetic scenarios were created to simulate possible intersections between a unified system-obstacle ellipsoid approximation and the actuation plane $\pplane \in \mathbb{R}^2$. This resulted in $100$ ellipses parameterised with semi-axis values ${\vprojectedgeometry=(\projectedwidth, \projectedheight)}$ uniformly sampled in the range $2.5$ to $25$cm. Each of these sections was placed in the middle of a one-metre length start-goal baseline. For each scenario, a set of trajectories were generated using \eref{eq:oa_complete} with a ${50\times50\times50}$ grid of the parameters ${\{\newgamma, \newbeta, \newkappa\}}$. Only those input-target ${\{(\projectedwidth, \projectedheight),(\newgamma, \newbeta, \newkappa)\}}$ pairs involving a collision-free trajectory were integrated into the dataset along with the resulting clearance.

        The \ac{RC} architectures were trained using a $70\%$ of the synthetic dataset. Each \ac{NN} was trained independently using the Levenberg-Marquardt algorithm with a random initialisation of the weights and biases. The remaining $30\%$ of the dataset was used to test the performance of the trained \ac{RC} models. Since the aim of a \ac{RC} model is to reduce the prediction error on every single target~\cite{spyromitros2016multi}, each model $Y_i$ was validated by computing the \ac{NMSE} on the training and testing sets. As shown in \tref{tab:nn_training}, the parameter prediction error of the models reduces significantly when considering the clearance in the input vector $\inputvector$. This is because the clearance allows differentiating the influence of the targets among all possible collision-free trajectories. It is worth noting that the performance of the \ac{RC} does not deteriorate when being evaluated on the test set.

    \subsection{Experiments on Familiar Scenarios}
\begin{figure}[t]
            \vspace{-0.3cm}
            \centering
            \subfloat[]{\includegraphics[height=5cm]{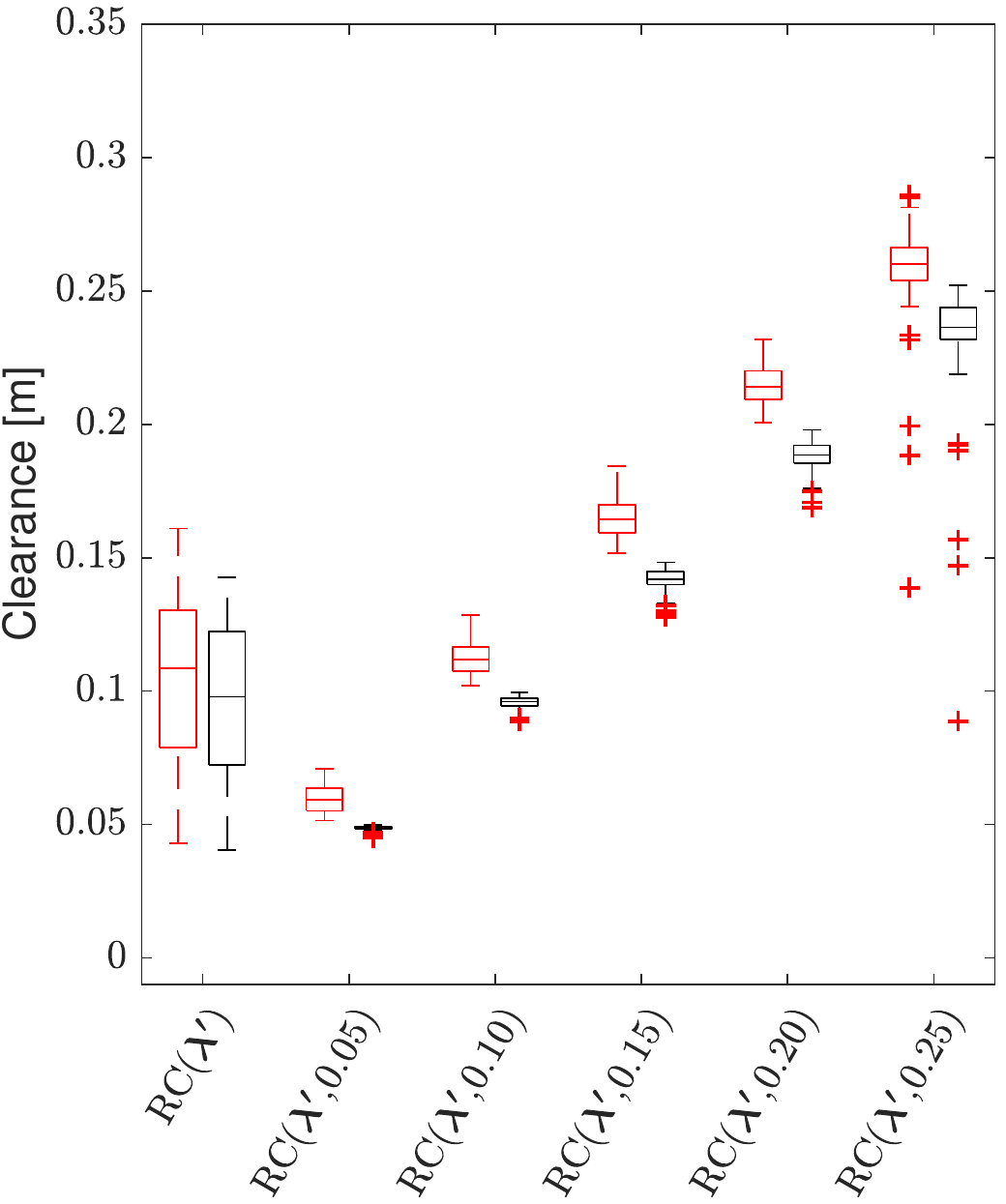} \label{fig:results_similar_clearance}}
            \subfloat[]{\includegraphics[height=5cm]{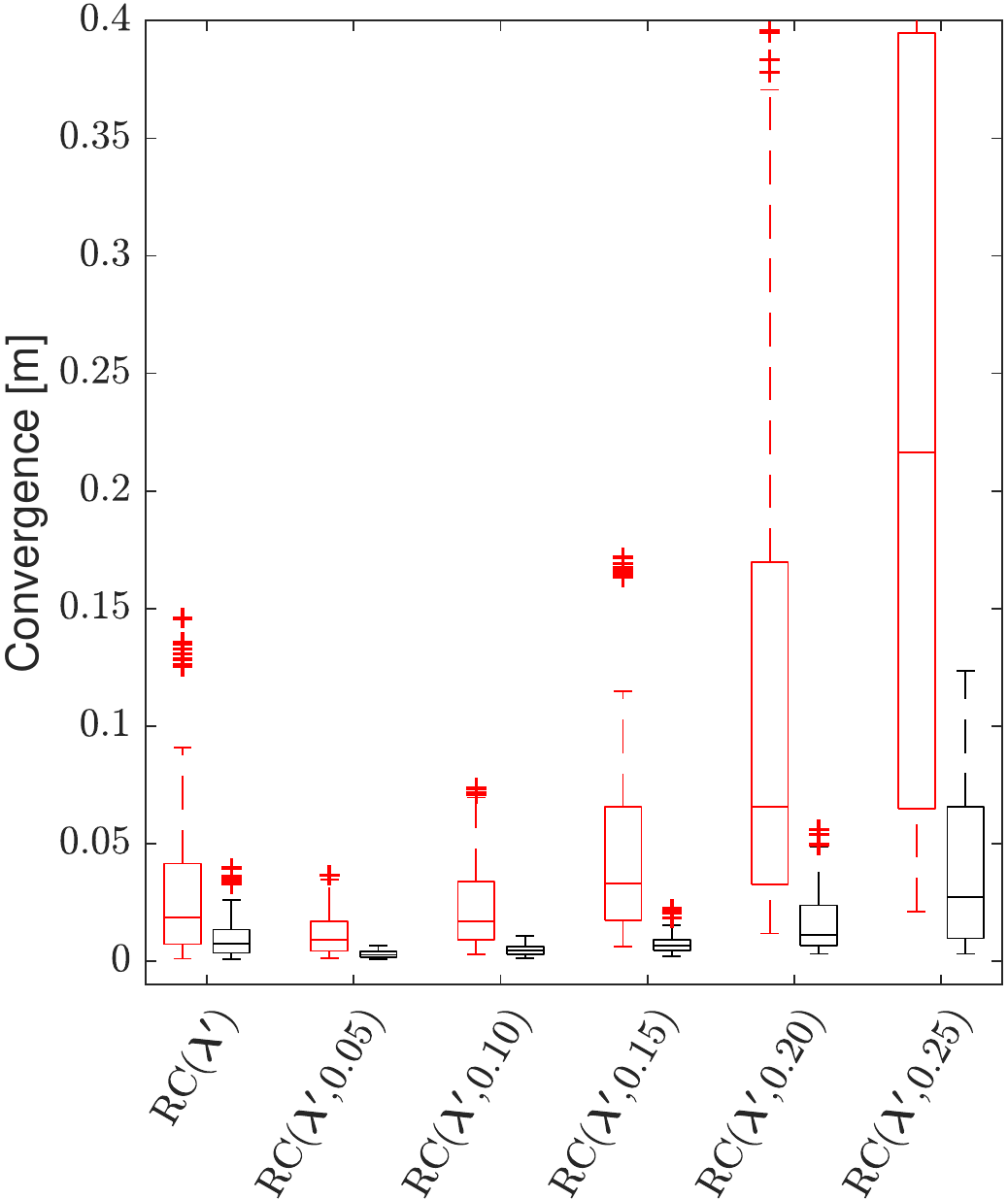} \label{fig:results_similar_convergence}}
            \caption{(a)~Clearance and (b)~convergence of the avoidance behaviours generated in familiar scenarios, when scaling the trajectory duration (black plots) and when not (red plots).}
            \label{fig:results_similar}
            \vspace{-0.3cm}
        \end{figure}

        The performance of both \ac{RC}($\vprojectedgeometry$) and \ac{RC}($\vprojectedgeometry,\clearance$) models in the $\pplane$-plane space was evaluated for the same obstacle geometries as in the training dataset, i.e. $100$ ellipses. For the \ac{RC}($\vprojectedgeometry,\clearance$) model, the considered constraints on the clearance were $\clearance=\{0.05, 0.1, 0.15, 0.2, 0.25\}$ metres. All six models were evaluated with and without scaling the trajectory duration $\tau$ according to its estimated length as explained in \sref{sec:geometry_route_goal_convergence}. Overall, this led to the testing of the \ac{RC} architecture under $12$ different settings. Performance in the $\pplane$-plane space was evaluated for the metrics (i)~number of collisions, (ii)~minimum distance to an obstacle (clearance), and (iii)~distance to goal (convergence). The obtained results over the $1{,}200$ scenarios are illustrated in~\fref{fig:results_similar}.

        \fref{fig:results_similar_clearance} and \fref{fig:results_similar_convergence} respectively represent the clearance to the obstacle and convergence to the goal for the $1{,}200$ scenarios evaluated across the $12$ settings of the \ac{RC} architecture.
        Overall, constraining the model with a desired clearance leads to more bounded behaviours. However, as the clearance constraint increases, the convergence rapidly deteriorates for those models not scaling the trajectory duration (red boxes). Instead, when scaling the time (black boxes), the convergence is at most of $3$cm for the most constrained model \ac{RC}($\vprojectedgeometry,0.25$). This fact highlights the importance of scaling the time when larger trajectories are required. Indifferently from the model setup, none of the $1{,}200$ conducted tests resulted with a trajectory colliding with an obstacle. The remainder of the experimental evaluation is conducted with the \ac{RC}($\vprojectedgeometry,0.15$) model and scaling the trajectory duration according to its estimated length.

    \subsection{Experiments on Novel Scenarios} \label{sec:results_unseen}
        Given the variety of obstacle avoidance scenarios that a system may face in the real-world, the proposed \ac{RC}($\vprojectedgeometry,0.15$) model was evaluated for its performance and generalisation capabilities on scenarios not seen during the training process. Notably, the approach was tested for its suitability to deal with \ac{3D} obstacles via the extraction of relevant unified low-dimensional geometrical features laying on the $\pplane$-plane as described in \sref{sec:geometry_sq}.

        Novel \ac{3D} scenarios were created by sampling the location and dilated geometry of the obstacle randomly. The obstacle was arbitrarily located along the x-axis between the start and goal configurations preserving $5$cm of margin, and around the baseline between $\minus0.4$ and $0.4$m along both the y-axis and z-axis.
        The unified system-obstacle ellipsoid approximation had random width, height and length within the spectrum $5$ to $50$cm, leading to representative candidates of possible object geometries in real-world environments. This spectrum corresponds to semi-axis values $\lambda_1$, $\lambda_2$ and $\lambda_3$ laying in the range $2.5$ to $25$cm. These boundaries also ensured that none of the extracted low-dimensional features $\vprojectedgeometry$ would result beyond the limits for which the \ac{RC} model was trained for.

\begin{figure}[b]
			\vspace{-0.2cm}
            \centering
            \subfloat[]{\includegraphics[height=2.88cm]{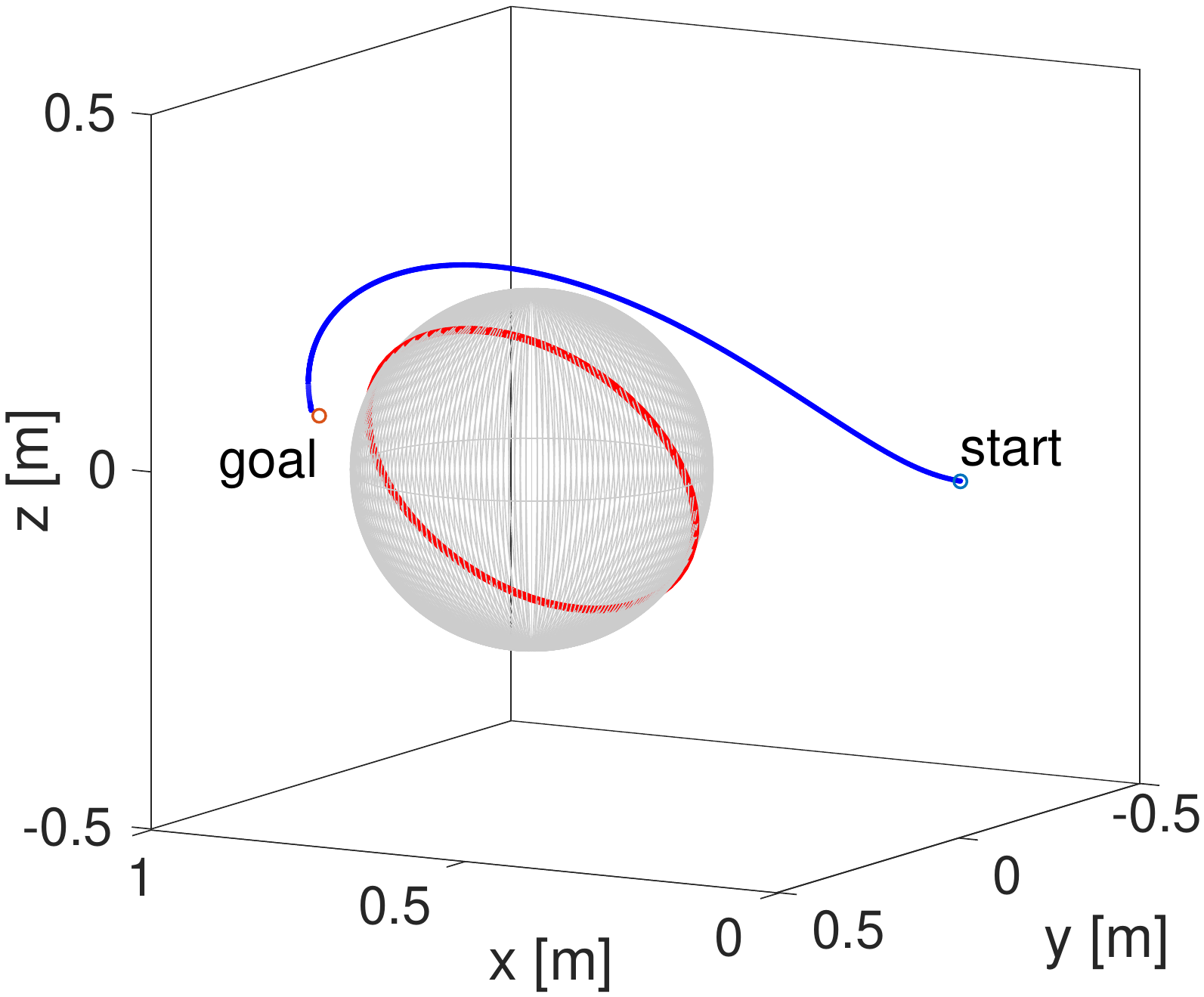} \label{fig:results_unseen_4}}
            \subfloat[]{\includegraphics[height=2.88cm]{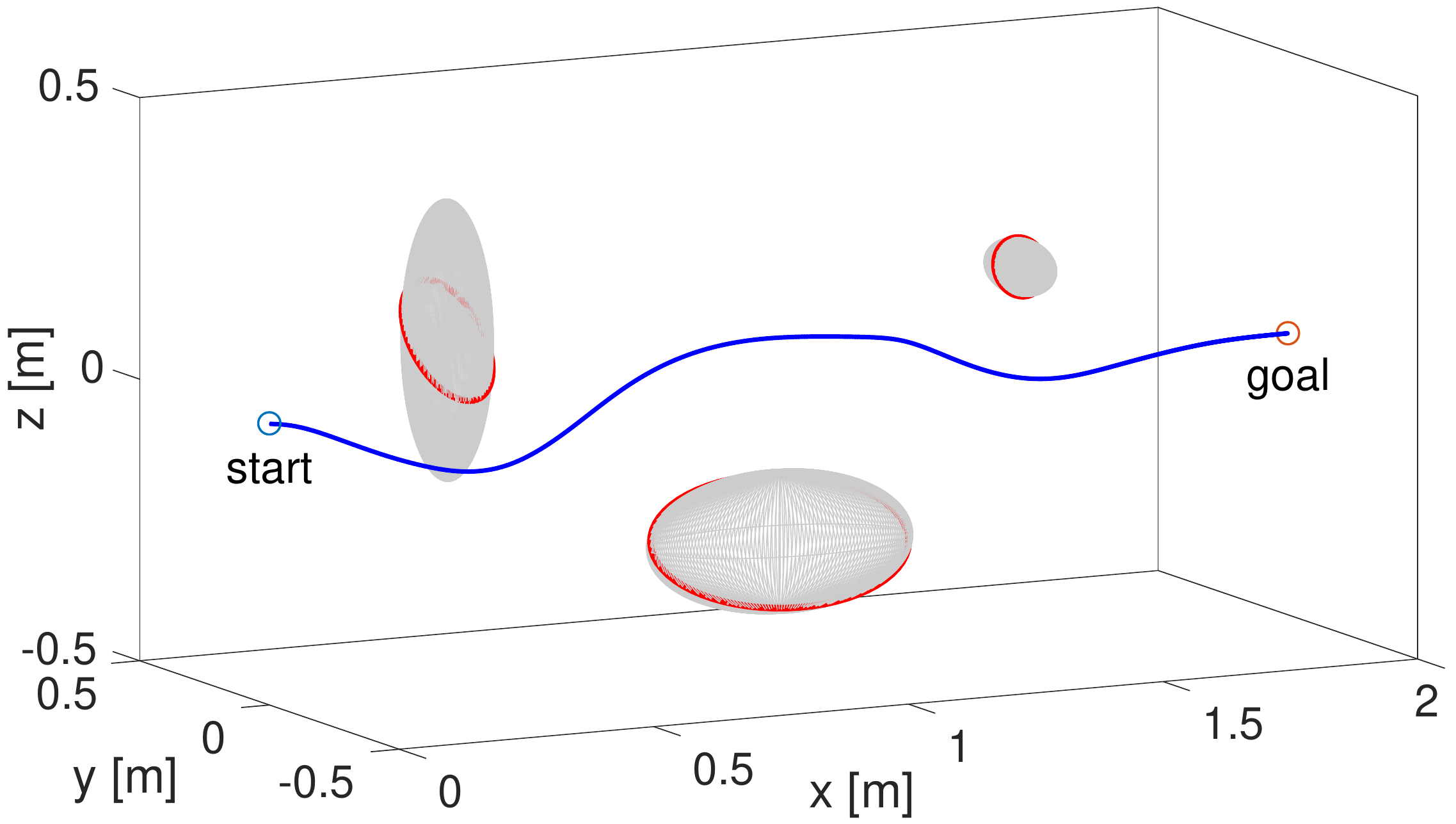} \label{fig:results_unseen_multi}}
            \caption{Generalisation capabilities of the trained \ac{RC}($\vprojectedgeometry,0.15$) model in novel settings. Parameters $\vprojectedgeometry$ are extracted from the relevant section (red ellipse) where the coupling term acts.}
            \label{fig:results_unseen}
        \end{figure}

        A $1{,}000$ novel \ac{3D} scenarios were created for the different start-to-goal baselines of $0.5$, $1.0$, $1.5$ and $2.0$m along the x-axis of the local frame, adding up to a total of $4{,}000$ evaluations. The semi-axis $\lambda_1$ was limited to a maximum of $20$cm for the baseline of $0.5$m to be consistent with the $5$cm margin across experiments. All environments required the action level to modulate a start-to-goal policy to avoid collision and preserve the desired clearance. Out of the $4{,}000$ tests, $1{,}296$ environments already had the baseline in collision with the obstacle. The performance of \ac{RC}($\vprojectedgeometry,0.15$) on the unseen settings was evaluated for the metrics (i)~number of collisions, (ii)~clearance to an obstacle, and (iii)~convergence to goal. \tref{tab:results_unseen_metrics} summarises the extracted metrics across the evaluation, and \fref{fig:results_unseen} depicts the performance of the proposal on some novel single and multi-obstacle settings.

        \begin{table}[t]
            \centering
            \begin{tabular}{lccccc}
                \toprule
                & \multicolumn{2}{c}{Clearance to} & \multicolumn{2}{c}{Convergence} & $n^\circ$ of \\
                & \multicolumn{2}{c}{obstacle [m]} & \multicolumn{2}{c}{to goal [m]} & collisions \\
                \cmidrule(lr){2-3} \cmidrule(lr){4-5}
                & mean & min & mean & max & \\
                \cmidrule{2-6}
                Goal at $0.5$m & 0.144 & -5.01e-4 & 4.41e-4 & 0.017 & 2 \\
                Goal at $1.0$m & 0.184 & 0.060 & 4.23e-4 & 0.017 & 0 \\
                Goal at $1.5$m & 0.196 & 0.068 & 5.22e-4 & 0.023 & 0 \\
                Goal at $2.0$m & 0.202 & 0.076 & 6.49e-4 & 0.027 & 0 \\
                \bottomrule
            \end{tabular}
            \caption{Clearance, convergence and number of collisions of the trained \ac{RC}($\vprojectedgeometry,0.15$) model for $4{,}000$ novel settings.}
            \label{tab:results_unseen_metrics}
			\vspace{-0.2cm}
        \end{table}

        Results in \tref{tab:results_unseen_metrics} reflect the performance of the designed \ac{RC}($\vprojectedgeometry,0.15$) model when dealing with \ac{3D} scenarios via their section on the $\pplane$-plane. The overall success rate is of $99.95\%$ on novel scenarios while providing, in average, a clearance similar to the requested one of $0.15$m and a close convergence to the goal. This implies an enhancement of $31.75$ times over the success rate reported on known objects in~\cite{rai2017learning}.
        However, the performance of the approach is slightly compromised in some scenarios, obtaining clearances of $6$cm and convergences up to $2.7$cm. The proposed approach could not cope uniquely with two scenarios out of $4{,}000$, where the generated trajectory penetrated $0.501$mm an obstacle of $40$cm along the x-axis and $50$cm along the y-axis and z-axis placed in the middle of a $0.5$m long baseline. Albeit these extreme scenarios for which more data could be provided at training time, the proposed approach has proved to generalise not only to different object sizes and locations, but also to different start-to-goal baselines. Further experimentation also showed the suitability of the framework to deal with multi-obstacle scenarios (see~\fref{fig:results_unseen_multi}). Since the action level is referenced in a local frame (see~\sref{sec:background_dmp}), the performance of the framework does not deteriorate regardless of the local frame's pose in the task space. Within the local frame, the outstanding generalisation capabilities are mainly due to regulating action according to the relative system-obstacle state defining the $\pplane$-plane, and extracting relevant system-obstacle low-dimensional geometrical descriptors.

    \subsection{Experiments on a Robotic Platform}

        \begin{figure*}[t]
			\centering
			\subfloat[]{\includegraphics[height=4cm]{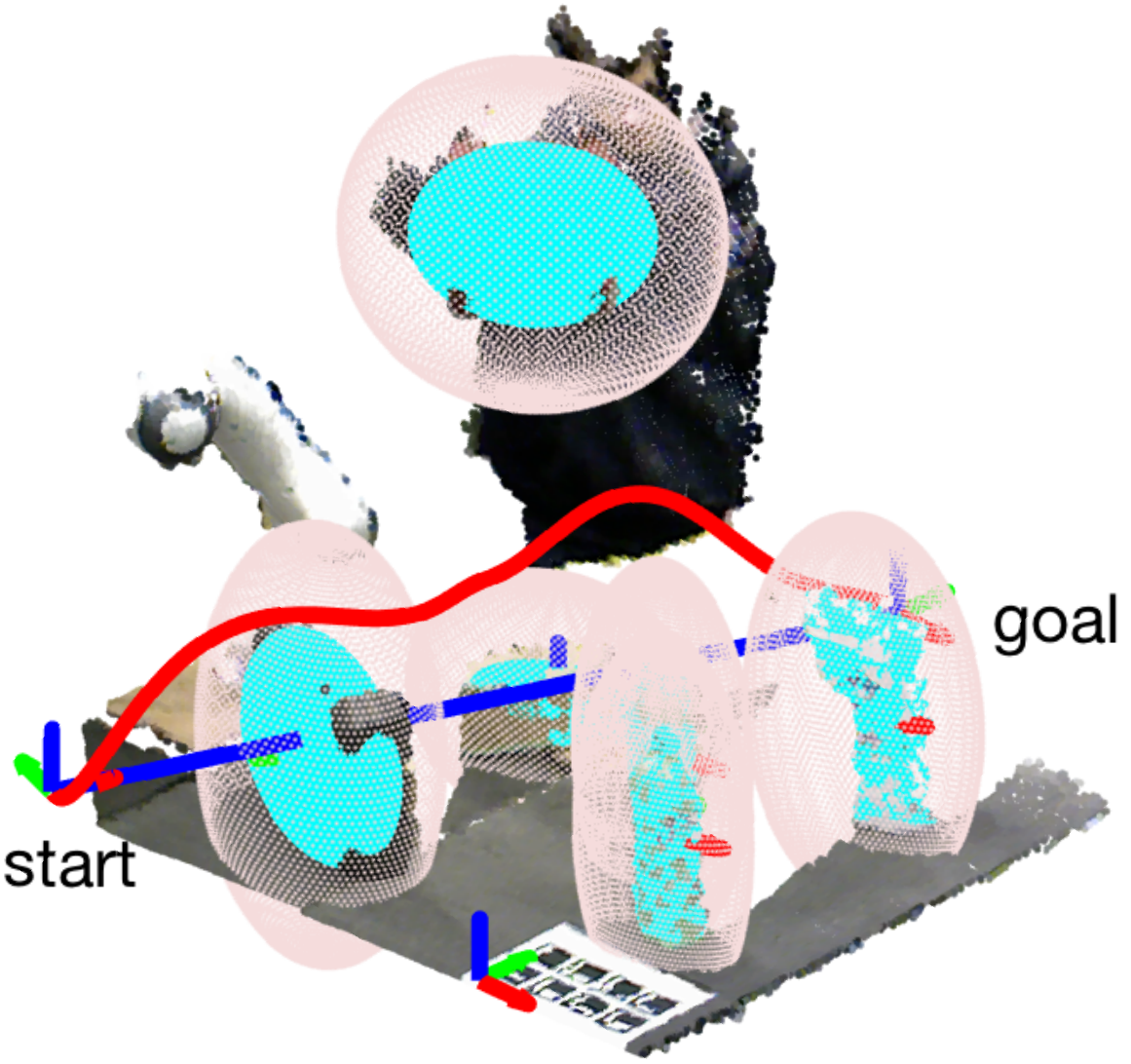}\label{fig:franka_perception}}
			\qquad
			\subfloat[]{\includegraphics[height=4cm]{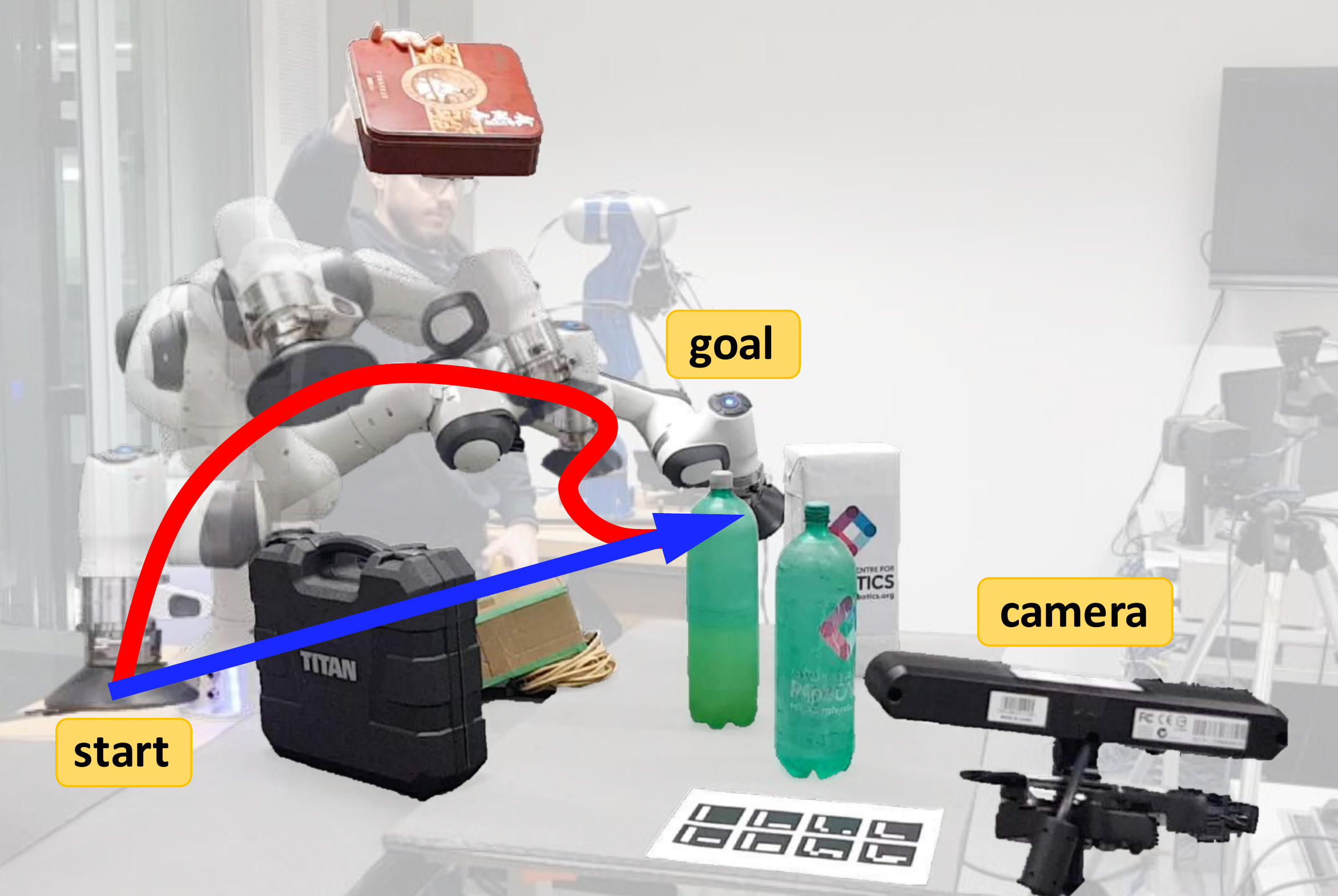}\label{fig:franka_exp_multi}}
			\qquad
			\subfloat[]{\includegraphics[height=4cm]{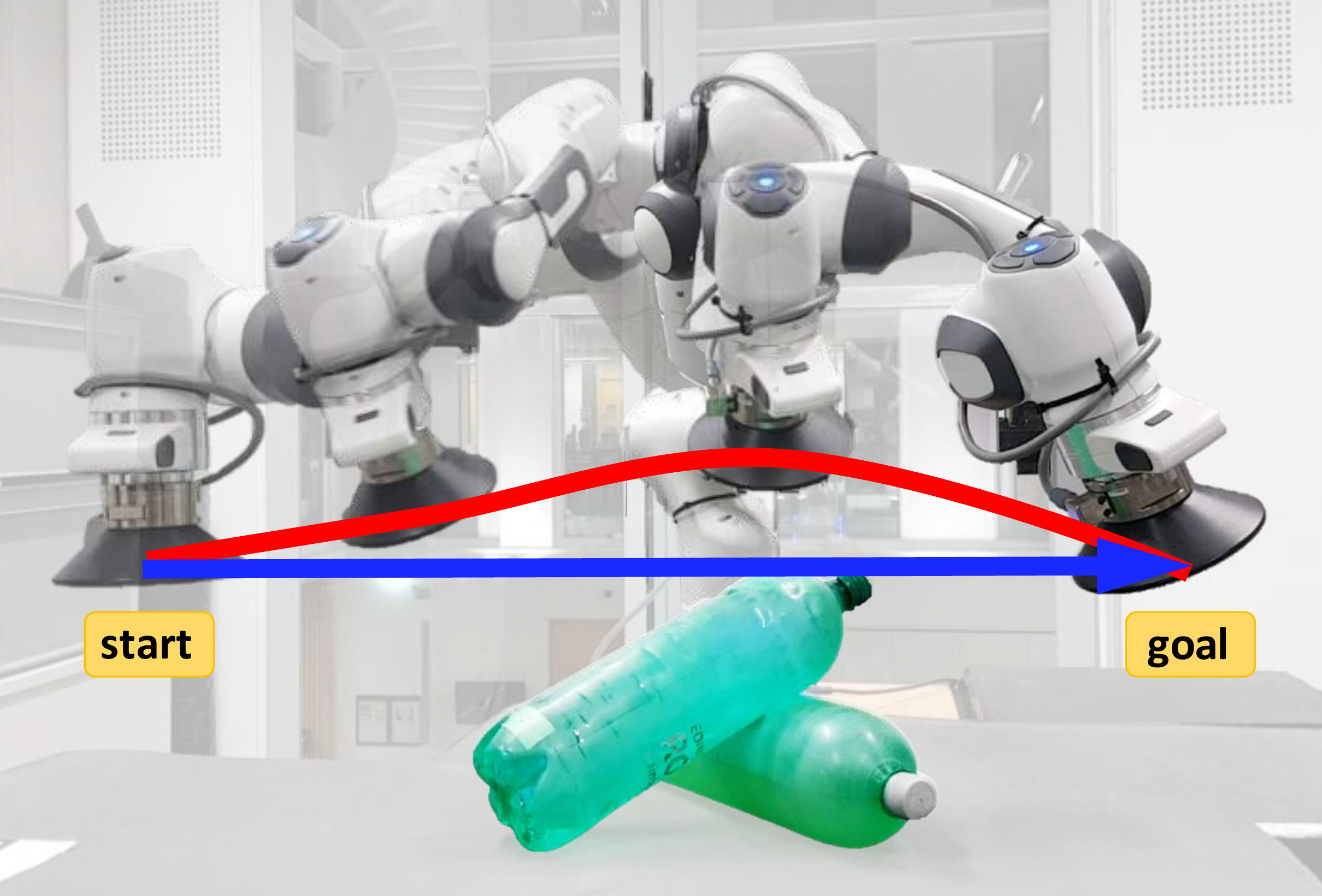}\label{fig:franka_exp_irregular}}
			\caption{Panda arm engaged in a start-to-goal policy (blue trajectories) while modulating its behaviour (red trajectories). (a)~Environment perception with unified low-dimensional encoding of the system's and obstacle's geometry (rose ellipsoids). Proposed hierarchical framework dealing with (b)~multiple obstacles in a cluttered environment, and (c)~an irregular obstacle.}
	        \label{fig:franka_experiments}
	    \end{figure*}

        \begin{table}[b]
            \vspace{-0.3cm}
            \centering
            \begin{tabular}{ccccc}
                \toprule
                & & \multirow{2}{*}{\shortstack{Clearance to \\ obstacle [m]}} & \multirow{2}{*}{\shortstack{Convergence \\ to goal [m]}} & \multirow{2}{*}{\shortstack{$n^\circ$ collisions $\&$ \\ $\clearance < 0.15$m}} \\
                & & & & \\
                \cmidrule{3-5}
                regular & pp. & fail & fail & 1 $\&$ 1 \\
                (Fig.~\ref{fig:intro}) & mod. & 0.168 & 9.58e-3 & 0 $\&$ 0 \\
                \cmidrule{2-5}
                irregular & pp.& fail & fail & 1 $\&$ 1 \\
                (Fig.~\ref{fig:franka_exp_irregular}) & mod. & 0.163 & 4.17e-3 & 0 $\&$ 0 \\
                \cmidrule{2-5}
                multi-obs & pp. & fail & fail & 2 $\&$ 5 \\
                (Fig.~\ref{fig:franka_exp_multi}) & mod. & 0.172 & 1.29e-2 & 0 $\&$ 0 \\
                \bottomrule
            \end{tabular}
            \caption{Clearance, convergence, number of collisions and unsatisfied clearances for the pre-planned (pp.) and modulated (mod.) start-to-goal policies on the real robot.}
            \label{tab:franka_metrics}
        \end{table}

        The proposed hierarchical framework for obstacle avoidance has been deployed on an anthropomorphic 7-\ac{DoF} Franka Emika Panda arm operated with \acs{OROCOS}~\cite{bruyninckx2001open}. The \ac{DMP}-encoded system's transient behaviour is converted to joint configurations using a Cartesian inverse dynamic controller with null space optimisation. 
        The environment is partially observed with a depth camera ASUS Xtion previously calibrated with Aruco markers~\cite{garrido2014automatic}. The acquired point cloud is processed applying standard filtering techniques to segment the clusters describing obstacles and the table. The partial observation of each obstacle is dilated to also account for the system's geometry (see \sref{sec:geometry_unified}). The location of the table is used to constrain the reactive behaviour along the upper part of the task space (see \sref{sec:geometry_route_selection}).



        As in~\cite{rai2014learning,rai2017learning}, the test-bed consisted of obstacles interrupting a straight trajectory underlying a start-to-goal policy. However, differently than~\cite{rai2014learning,rai2017learning}, the assortment of considered obstacles had not been seen before. This included, but was not limited to, regular objects, such as the cardboard box in \fref{fig:intro}, irregular objects, such as the pile of plastic bottles in \fref{fig:franka_exp_irregular}, and also aleatory combinations of them, such as the cluttered environment with six obstacles in \fref{fig:franka_perception} and \fref{fig:franka_exp_multi}. As summarised in \tref{tab:franka_metrics}, the robot engaged in the pre-planned policy (blue trajectories) would impact with the obstacles. Instead, endowing the robot with the ability to modulate such policy, allows the system to successfully circumvent all obstacles with the desired $0.15$m clearance while converging to the goal (red trajectories).

	    The presented results demonstrate that the proposed hierarchical framework which (i)~extracts relevant geometric descriptors, (ii)~uses them in the designed \ac{RC}-based learning module to (iii)~regulate the \ac{DMP}-based action level, endows a system with the ability to modulate its behaviour in settings never seen before, while stably converging to the goal.

    \section{FINAL REMARKS} \label{sec:conclusions}
	
	This paper has presented a biologically-inspired hierarchical framework which safely modulates an on-going policy to avoid obstacles. The proposed approach follows a multi-layered perception-decision-action analysis which (i)~extracts unified system-obstacle low-dimensional geometric descriptors, then (ii)~exploits them to rapidly reason about the environment with a combination of heuristics and learning techniques, and finally (iii)~guides and regulates the obstacle avoidance behaviour with a conjunction of coupling terms modulating a \ac{DMP}-encoded policy. Experimentation conducted in synthetic environments highlights this method's generalisation capabilities to confront novel scenarios at the same time of ensuring the convergence of the system to the goal. 
	Additionally, real-world trials on an anthropomorphic manipulator demonstrated the framework's suitability to successfully modulate a policy in the presence of multiple novel obstacles described by partial visual-depth observations, while satisfying a user-defined clearance constraint.
	
	
	The proposed framework is not restricted to the presented experimental evaluation nor platform. Any robotic system following a \ac{DMP}-encoded policy can benefit from this work to safely modulate its behaviour in the presence of unexpected obstacles. Similarly to~\cite{park2008movement}, collisions of the links can also be considered by finding the closest geometric section on the robot to the obstacle, and then modulating the kinematic null-space movement with the proposed approach. An interesting venue for future work is to modulate the system's orientation policy to overcome an obstacle, which, for instance, might have a significant impact on a manipulator carrying a large bulk. Another interesting extension of this work is learning route selection priorities in cluttered environments, so systems can autonomously reason about the most convenient direction to avoid an obstacle.


	\bibliographystyle{ieeetr}
    \bibliography{refs.bbl}
\end{document}